\documentclass[runningheads]{llncs}

\usepackage{eccv}

\usepackage{eccvabbrv}

\usepackage{graphicx}
\usepackage{booktabs}

\usepackage[accsupp]{axessibility}  % Improves PDF readability for those with disabilities.

\usepackage{hyperref}

\usepackage{orcidlink}

\usepackage[utf8]{inputenc} % allow utf-8 input
\usepackage[T1]{fontenc}    % use 8-bit T1 fonts
\usepackage{url}            % simple URL typesetting
\usepackage{booktabs}       % professional-quality tables
\usepackage{amsfonts}       % blackboard math symbols
\usepackage{nicefrac}       % compact symbols for 1/2, etc.
\usepackage{microtype}      % microtypography
\usepackage{graphicx}
\usepackage{graphicx}
\usepackage{amssymb}
\usepackage{amsmath}
\usepackage{witharrows}
\usepackage{makecell}
\usepackage{multirow}
\usepackage{color}
\usepackage{chemformula}
\usepackage{floatrow}
\usepackage{tikz}
\usepackage{algorithm}  
\usepackage{algpseudocode}  
\usepackage{algorithmicx}  
\usepackage{colortbl}
\usepackage{floatrow}
\usepackage{textcomp}

\usepackage{array, caption, floatrow, tabularx, makecell, booktabs}%

\usetikzlibrary {arrows.meta,bending,positioning}

\begin{document}

\title{Make a Strong Teacher with Label Assistance:\\A Novel Knowledge Distillation Approach \\ for Semantic Segmentation} 

\titlerunning{Label Assisted Distillation}

\author{Shoumeng Qiu\inst{1}\orcidlink{0000-0003-4475-2303} \and
Jie Chen\inst{1}\orcidlink{0000-0002-5625-5729} \and
Xinrun Li\inst{2}\orcidlink{0000-0002-2548-2187} \and
Ru Wan\inst{3}\orcidlink{0009-0008-8151-0059}\and
Xiangyang Xue\inst{1}\orcidlink{0000-0002-4897-9209}\and
Jian Pu\inst{1}*\orcidlink{0000-0002-0892-1213}}


\institute{Fudan University, Shanghai, China
\\
\email{smqiu21@m.fudan.edu.cn; \{chenji19,xyxue,jianpu\}@fudan.edu.cn}
\and Bosch Corporate Research, China \and
Mogo.ai Information and Technology Co., Ltd, Beijing, China \\
\email{wanru@zhidaoauto.com}
}

\maketitle

\begin{abstract}
  In this paper, we introduce a novel knowledge distillation approach for the semantic segmentation task. Unlike previous methods that rely on power-trained teachers or other modalities to provide additional knowledge, our approach does not require complex teacher models or information from extra sensors. Specifically, for the teacher model training, we propose to noise the label and then incorporate it into input to effectively boost the lightweight teacher performance. To ensure the robustness of the teacher model against the introduced noise, we propose a dual-path consistency training strategy featuring a distance loss between the outputs of two paths. For the student model training, we keep it consistent with the standard distillation for simplicity. Our approach not only boosts the efficacy of knowledge distillation but also increases the flexibility in selecting teacher and student models. To demonstrate the advantages of our Label Assisted Distillation (LAD) method, we conduct extensive experiments on five challenging datasets including Cityscapes, ADE20K, PASCAL-VOC, COCO-Stuff 10K, and COCO-Stuff 164K, five popular models: FCN, PSPNet, DeepLabV3, STDC, and OCRNet, and results show the effectiveness and generalization of our approach. We posit that incorporating labels into the input, as demonstrated in our work, will provide valuable insights into related fields. Code is available at \url{https://github.com/skyshoumeng/Label_Assisted_Distillation.}
  
  \keywords{Semantic Segmentation \and Knowledge Distillation \and Privileged Information}
\end{abstract}

\section{Introduction}
\label{sec:introduction}

\begin{figure}[t]
\centering
\includegraphics[width=.85\linewidth]{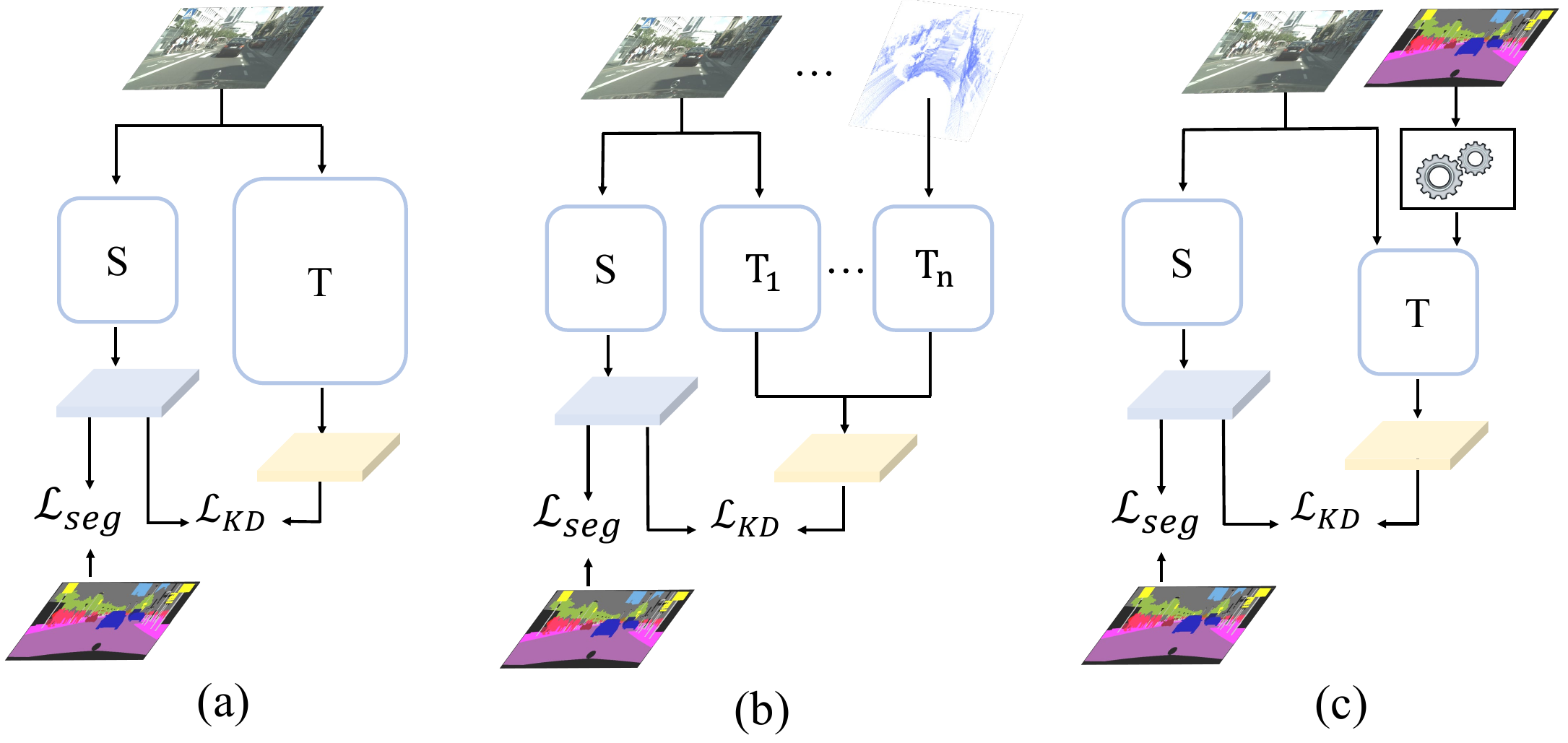}
\caption{Comparison with two main categories of distillation approaches. S and T indicate the student and teacher model, $\mathcal{L}_{seg}$ denotes the segmentation task loss and $\mathcal{L}_{kd}$ denotes the distillation loss. Additional knowledge mainly from:  (a) power-pretrained teacher model; (b) teacher that takes additional modalities as input; (c) lightweight teacher model that takes processed label as input. Our approach (c) shows a clear advantage as it has no requirements for complex teacher model or other modalities.
}
\label{fig1:comp}
\end{figure}
Semantic segmentation is one of the most fundamental tasks that aims to classify every pixel of a given image into a specific class. It is widely applied to many applications such as autonomous driving \cite{siam2018comparative,xie2021segformer,wang2022rtformer}, video surveillance \cite{zhang2021video,heo2022vita}, and biomedical image diagnosis \cite{ronneberger2015u,alzahrani2021biomedical}. Nevertheless, most of the existing models entail high complexity and heavy computational cost for achieving superior performance \cite{xiao2018unified,he2019adaptive,zheng2021rethinking,xie2021segformer}, which is inappropriate in real-world applications. To alleviate these issues, knowledge distillation\cite{hinton2015distilling} has been introduced into segmentation \cite{he2019knowledge,chen2021cross,vobecky2022drive,liu20213d}, which aims at transferring knowledge from more complex powerful teacher models to efficient lightweight student models. 

Based on the source of additional knowledge primarily deriving from the teacher model, current knowledge distillation methods for segmentation broadly fall into two categories: knowledge from the more powerful model capacity, such as \cite{he2019knowledge,chen2021cross,shu2021channel}, and knowledge from extra data information, such as \cite{vobecky2022drive,liu20213d,yan20222dpass}. 
However, the approaches in both categories have some apparent issues. 
For the first category, where additional knowledge is derived from the teacher model, this approach typically requires using a more complex and computationally expensive model as the teacher to extract more useful information from the inputs. \cite{liu2020structured,shu2021channel,wang2020intra,yang2022cross}. For the second category, where additional knowledge comes from extra data information, this always involves other modalities \cite{hu2020knowledge,yan20222dpass} such as infrared or LiDAR data which are often costly or unattainable in practice. For the above issues, we aim to conduct knowledge distillation learning using a lightweight teacher model and eliminate the need for additional modalities. 

In this paper, we innovatively consider improving the performance of lightweight teacher model by incorporating label information into the input, as the label is always attainable in supervised learning tasks. The key difference between the existing method and our proposed approach is shown in Fig. \ref{fig1:comp}. Specifically, for the teacher model, unlike previous methods that adopt complex models or extra information to achieve higher performance, we improve the performance by taking the label as part of the input to the model.
However, there is a problem with directly feeding labels into the model, as the model may take shortcuts in mapping inputs to outputs based on the labels rather than learning the intended solution as mentioned in \cite{geirhos2020shortcut,scimeca2021shortcut}. To address this, we propose to noise the label randomly before feeding it into the model. In experiments, we found that there are fluctuations in the output of the teacher model due to the random sampling in the label noising operation. To counteract the effects of the introduced noise, we further propose a dual-path consistency training strategy with a consistency loss between the outputs of two paths. When the teacher model is trained, the distillation learning for the student model is the same as the standard distillation approach. It should be noted that the noised label in this paper is different from \cite{li2017learning,das2023understanding,lu2016learning,li2016large}, where label noise originates from annotations, our approach intentionally introduces noise to the clean label.

In addition, our approach allows for a more flexibility choice of teacher and student models. For example, the teacher can be more complex than the student, the same as the student, or even simpler than the student. 
Finally, to evaluate the effectiveness and generalization of our approach, we conduct extensive experiments on five baseline segmentation models: FCN \cite{long2015fully}, PSPNet \cite{zhao2017pyramid}, DeeplabV3 \cite{chen2017rethinking}, STDC \cite{fan2021rethinking}, and OCRNet \cite{yuan2020object}, and across five challenging datasets: Cityscapes \cite{cordts2016cityscapes}, ADE20K \cite{zhou2017scene}, PASCAL-VOC 2012 \cite{everingham2010pascal}, COCO-Stuff 10K \cite{caesar2018coco}, and COCO-Stuff 164K \cite{caesar2018coco}.  The experimental results show consistent performance improvement across all cases. We also perform detailed analyses of crucial components in the proposed approach and hope other researchers can gain inspiration from our study. Our contributions are summarized as follows:

\begin{itemize}

\item We propose a novel knowledge distillation approach with noised labels as privileged information for the semantic segmentation task. Our approach alleviates the dependency on complex teacher models or other modalities, and can effectively improve performance of knowledge distillation.

\item To enhance the robustness of teacher against the noise introduced in privileged information, we propose a dual-path consistency training strategy with a distance loss to minimize discrepancies between the outputs of two paths.

\item We conduct extensive experiments on five popular semantic segmentation baseline models across five challenging public datasets, and experimental results show substantial and consistent improvements on performance. We also perform analyses on crucial components of the proposed approach.

\end{itemize}

\section{Related works}

\subsection{Semantic Segmentation}

Semantic segmentation is one of the most fundamental computer vision tasks, which aims at classing every pixel on an input image into a certain class. It has a wide range of applications, such as autonomous driving and video surveillance. Deep neural network-based approaches are dominant in this task. FCN \cite{long2015fully} is a fully convolutional model that can take input of arbitrary size and produce correspondingly-sized output with efficient inference and training. PSPNet \cite{zhao2017pyramid} proposed to exploit the capability of global context information by different-region-based context aggregation operation through the pyramid pooling module and the pyramid scene parsing network. DeeplabV3 \cite{chen2017rethinking} proposed to combine both the spatial pyramid pooling module and encode-decoder structure. In this way, the approach can have the advantage of multi-scale contextual information encoding and sharper object boundary capturing at the same time. In recent years, many computational and memory-intensive models \cite{zheng2021rethinking,strudel2021segmenter,xie2021segformer} have been proposed to further improve the performance on this task, However, these models are not friendly to applications in real-world scenarios. In response to this problem, many real-time methods have been proposed \cite{li2019dfanet,yu2018bisenet,fan2021rethinking}. For example, STDC \cite{fan2021rethinking} proposed an efficient Short-Term Dense Concatenate network structure by gradually reducing the dimension of feature maps and using the aggregation of them for better image representation.

\subsection{Knowledge Distillation}
% In this paper, we mainly focus on the distillation approaches applied to the semantic segmentation task. 
Knowledge distillation is a technique for model compression and acceleration in which a smaller model, referred to as the student model, is trained to mimic the behavior of another model, usually a larger pre-trained model, known as the teacher model. It is popularized by \cite{hinton2015distilling} and has attracted a surge of attention in recent years \cite{nguyen2021dataset,wang2021knowledge,stanton2021does}. Surveys \cite{wang2021knowledge} and \cite{hu2023teacher} summarized methods from various perspectives, and our method can be categorized into Cross-Modal distillation (multi-modal to single modal). The key distinction between our approach and previous methods is our innovative use of noised labels as an input modality. 

Based on the source of additional knowledge primarily comes from in the teacher model. We roughly categorize current works on knowledge distillation into two main categories: knowledge from a more complex teacher model and knowledge from extra information. Here we mainly focus on the distillation approaches applied to the semantic segmentation task. For the knowledge from a more complex teacher model, there is currently a lot of work devoted to further improving the efficiency and performance of the knowledge distillation approach, such as SKD\cite{liu2020structured}, IFDV\cite{wang2020intra}, CWD\cite{shu2021channel}, CIRKD\cite{yang2022cross}. SKD \cite{liu2020structured} proposed to distill structured knowledge from the teacher model to the student model as dense prediction is a structured prediction problem. IFDV \cite{wang2020intra} proposed an Inter-class Distance Distillation method to transfer the inter-class distance in the feature space from the teacher network to the student network. CWD \cite{shu2021channel} introduced to minimize the Kullback–Leibler (KL) divergence between the channel-wise probability map of the two networks. CIRKD\cite{liu2020structured} proposed to transfer structured pixel-to-pixel and pixel-to-region relationships across entire images. For the knowledge from extra information, LGD \cite{zhang2022lgd} and LG3D \cite{huang2022label} introduced distillation with label methods involving manually designed label encoder and label mapper modules, which specifically designed for object detection task, different from the above methods, our proposed approach is label encoder and label mapper free for the semantic segmentation task. KD-Net \cite{hu2020knowledge} proposed a framework to transfer knowledge from multi-modal to a mono-modal for medical image segmentation. 2DPASS \cite{yan20222dpass} proposed to distill richer semantic and structural information from the multi-modal data to boost the feature learning from single-modal data. 

\subsection{Learning using Privileged Information}

Learning using privileged information \cite{vapnik2009new} is also a crucial technique that enables machines to learn from other machines. The framework of learning using privileged information aims to leverage the additional information at training time to help boost the performance of the model, which is not accessible at the test time. In \cite{lopez2015unifying}, the authors proposed to unify the knowledge distillation and privileged information two into generalized distillation for machines learning from other machines. In \cite{feyereisl2012privileged}, the authors proposed to translate the notion of privileged information to the unsupervised setting in order to improve clustering performance. PISR\cite{lee2020learning} proposed to leverage the high-resolution (HR) images as privileged information and transfer the important knowledge of the HR images to a student model.

\begin{figure*}[t]
\centering
\includegraphics[width=1.\linewidth]{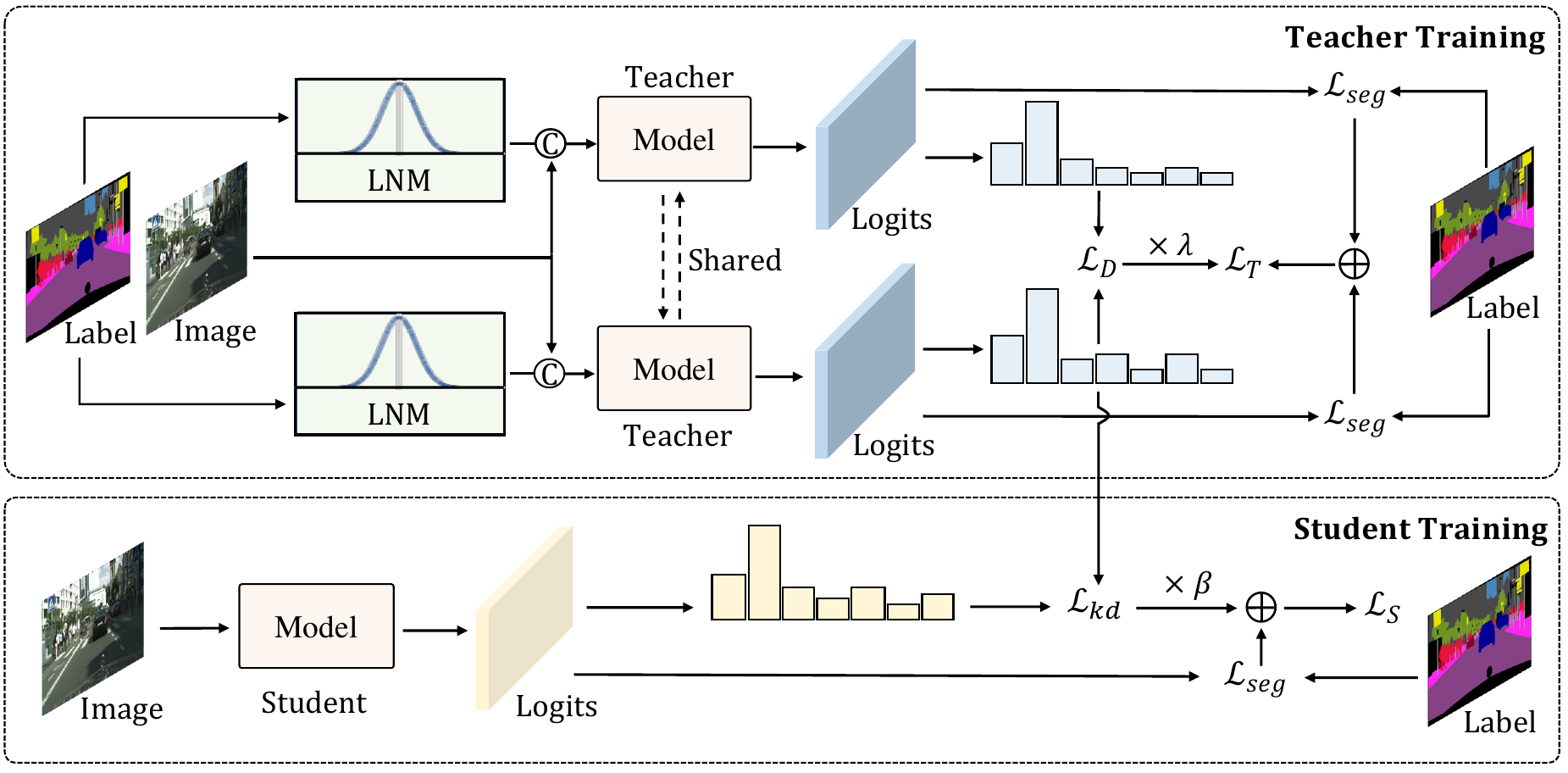}
\caption{The overall pipeline of our distillation framework. \textcopyright\;denotes the concatenation operation on channel, $\bigoplus$ denotes the addition operation. For the teacher model training within the dashed box above, the Label Noising Module (LNM) and the teacher model are duplicated into two copies. Then, the image and noised label are fed in respectively, The outputs of models are undergo supervision of the label. Additionally, we integrate a consistency constraint loss between the two predictions. We only need to retain one branch of teacher training for student model in distillation learning.}
\label{fig1:framework}
\end{figure*}

\section{Proposed Method}

The overall framework of our proposed distillation approach is shown in Fig. \ref{fig1:framework}. The main distinction between our framework and the conventional distillation framework lies in the input information and training strategy of the teacher model. We do not rely on power-pretrained teacher models \cite{liu2020structured,wang2020intra} or extra information coming from other modalities \cite{hu2020knowledge,yan20222dpass}, which gives our approach obvious advantages over other distillation methods.

\subsection{Problem Formulation}
\label{sec:probformu}
We consider improving the performance of lightweight teacher model by incorporating more task-related information into the input. However, as additional information is often unattainable or costly in practice, so we wonder if we could use the readily available information to achieve this purpose. Specifically, in this paper, we aim to introduce the label to the input of teacher model, which is always attainable in the supervised learning tasks.

The simplest approach is to directly incorporate the label into the model's input, then the objective of model learning changes from
\begin{equation}
\label{con:SL}
\{(X_1,Y_1),...,(X_n,Y_n) \} \sim P^n(X,Y), %label \gets {input},
\end{equation}
to 
\begin{equation}
\label{con:SL}
\{(X_1,Y_1),...,(X_n,Y_n) \} \sim P^n((X,Y),Y), %label \gets  \{input, \enspace label\} \; ,
\end{equation}
where each $(X_i, Y_i)$ is a image-label pair, $X_i \in \mathbb{R}^{H\times W \times 3}$, $Y_i \in \mathbb{R}^{H\times W}$ and $P(\cdot,\cdot)$ denotes the joint probability distribution. However, there is a problem with the above simple solution. Since $Y_i$ in the input and the expected output $Y_i$ are the same, the model may take a shortcut instead of learning the
intended solution \cite{geirhos2020shortcut,scimeca2021shortcut}, which means that the deep neural network may learn a mapping from the input label to the output rather than extract some useful features in the image, as indicated in  Eq. \ref{equ:SL}. Please refer to \emph{Shortcut Learning with Clean Label Input} in subsection \ref{ana:mainab} for more detailed discussions.
\begin{equation}
\begin{aligned}
\label{equ:SL}
\!\!\!\!\!\!\!\!\!\!\!\!\!\!\!\!\!\!
\centering
\vcenter{\hbox{
\begin{tikzpicture}
  \node (A)              {$\{X_i,\enspace Y_i\}$};
  \node (C) [node distance=.1cm, right=of A] {$  \longrightarrow \enspace $};
  \node (B) [node distance=.1cm, right=of C] {$Y_i$};
  \draw [->] --node[above=7.mm]{$\qquad\ \qquad \qquad shortcut $} (A) to [bend right=-40] (B);
\end{tikzpicture}
}},
\end{aligned}
\end{equation}
Since directly incorporating the label into the input is inappropriate, some transformation operations should be applied on the label before feeding it to the model. Here we denote the transformation operation as function $\phi(\cdot)$, then the objective of model learning become:
\begin{equation}
\begin{split}
\label{con:SL}
% & \{(X_1,\phi(Y_1),Y_1),...,(X_n,\phi(Y_n),Y_n) \}\\ 
% & \qquad \qquad \sim P^n((X,\phi(Y)),Y) %label \gets \{input, \enspace \phi(label) \},
 \{(X_1,\phi(Y_1),Y_1),...,(X_n,\phi(Y_n),Y_n) \}  \sim P^n((X,\phi(Y)),Y) %label \gets \{input, \enspace \phi(label) \} 
,
\end{split}
\end{equation}
For the function $\phi(\cdot)$, two conditions should be satisfied. First, based on the above analysis, it should make the mapping learning from $\phi(Y_i)$ to $Y_i$ difficult. Second, $\phi(Y_i)$ should maintain some useful information for the task, otherwise, $\phi(Y_i)$ will have no benefit for the predictions. In this paper, we propose a simple but effective Label Noising Module (LNM) as function $\phi(\cdot)$ which satisfies the above two conditions well.  We will give a detailed description to the implementation of LNM in the next subsection.

\begin{figure*}[t]
\centering
\includegraphics[width=1.\linewidth]{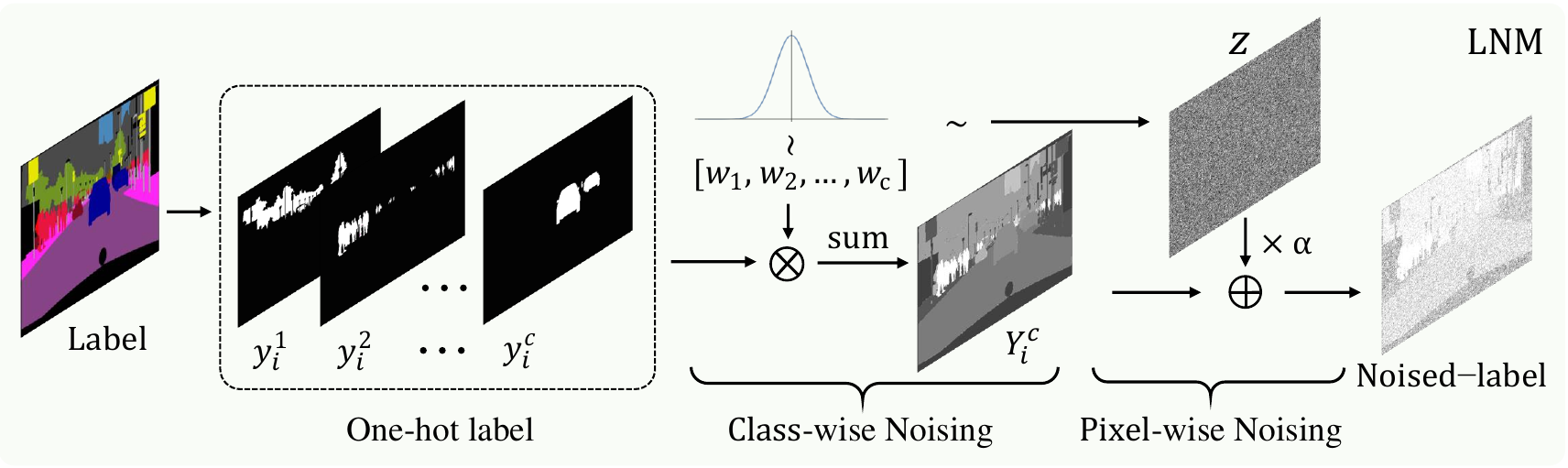}
\caption{Details of label noising module. $\bigotimes$ denotes the multiplication operation, $\bigoplus$ denotes the addition operation. The label is first represented in one-hot encoding, then the channel of each class is multiplied by a weight obtained from random sampling. After that, the results are added along the channel dimension, resulting in the distortion of the class index information. Subsequently, random noise is added for each pixel, yielding the final noised label, which is referred as \emph{privilege information} in this paper.
}
\label{fig1:teacher}
\end{figure*}

\subsection{Strong Teacher with Noised label}
\noindent \textbf{Design of LNM.} We note that the information contained in the label $Y_i$ can be decomposed into two parts: unique semantic class index and index consistency among pixels within the same class. Based on the above analysis, our proposed label noising module is also two-stage: class-wise noising and pixel-wise noising, respectively. For class-wise noising operation, we aim to distort the class index. We first represent the label $Y_i$ in the form of one-hot $Y_{i}^{oh} \in \{y_{1}^1, y_{i}^2, ..., y_{i}^c\}$, where $c$ is the total number of classes. For each $y_{i}^j \in \mathbb{R}^{H\times W}$, we perform a multiplication operation with $w_i$ which is sampled from a predefined distribution $\Omega_1$. Then we perform sum operation across channels and get the class-wise noising label $Y_i^c$. For pixel-wise noising operation, we aim to disrupt the index consistency in each class. We first generate a random matrix $Z \in \mathbb{R}^{H\times W} $ where each element is sampled from a predefined distribution $\Omega_2$, then we scaled $Z$ with $\alpha$ and add to the $Y_i^c$. Finally, the label noising operation can be expressed as follows:\begin{equation}
\begin{aligned}
\label{equ:trans}
\phi(Y_i) = sum(W \cdot Y_i^{oh}, \enspace dim\hspace{-0.1cm}=\hspace{-0.1cm}0) + \alpha Z \;,
\end{aligned}
\end{equation}
Here $W=\{w_1, w_2,…,w_c\}$. Up to now, the noised label is obtained. Then, we consider incorporating the noised label information into the input of model. We choose the simple concatenation operation as it is widely used in many multi-modal fusion works \cite{zhang2019multi,li2020enhanced}. Therefore, the teacher model takes a four-channel tensor as input, instead of the standard three RGB channels. We adopt the commonly used standard Gaussian distribution for both $\Omega_1$ and $\Omega_2$ in this paper. The pipeline of our label noising module is shown in Fig. \ref{fig1:teacher}.

Next, we give explanations about why the operation described in Eq. \ref{equ:trans} meets the two conditions mentioned in subsection \ref{sec:probformu}. First, since  $\phi(\cdot)$ employs random parameters for the noising operation, so there is no direct inverse function for $\phi(Y_i)$, which makes the mapping learning from $\phi(Y_i)$ to $Y_i$ difficult. Second, for the class-wise noising operation, as each $Y_i^{oh}$ shares the same parameter $w_i$, therefore, we can guarantee that the value within each class is still consistent after the class-wise nosing operation. For the pixel-wise noising operation, we can control the impact of this noise by adjusting parameter $\alpha$. Based on the above analysis, we can see that the generated noised label $\phi(Y_i)$ satisfies the conditions well and can be used as additional information to assist the model for better segmentation performance. 

%To minimize the distance between the outputs under two different sampling. 
\noindent \textbf{Dual-path Consistency Training.} In experiments, we found that there is a problem when adopting the noised label from Fig. \ref{fig1:teacher} for teacher training. Since parameters $W$ and $Z$ are sampled randomly in each training iteration, this will result in the fluctuation of the teacher output $O_i^1$ and $O_i^2$ with the same input image $X_i$ under two different sampled parameters $\theta_1$ and $\theta_2$ of LNM:
\begin{equation}
\begin{split}
 & \{X_i, \enspace \phi(Y_i|\theta_1) \} \rightarrow O_i^1, \enspace   \{X_i, \enspace \phi(Y_i| \theta_2) \} \rightarrow O_i^2,  \\
 & \qquad \qquad \qquad \mathcal{D}(O_i^1,\enspace  O_i^2)  >  0,
\end{split}
\end{equation}
here we adopt a more specific form for $\phi(\cdot)$ as $\phi(\cdot|\theta)$ for the sake of clarity, where $\theta_1$, $\theta_2$ containing both $W$ and $Z$.  $\mathcal{D}(\cdot,\cdot)$ is a function to measure the distribution distance between two outputs. In experiments, we found that the $\mathcal{D}(O_i^1, O_i^2)$ is a non-negligible value. This inconsistency makes it challenging for the student model to learn from the teacher, as the outputs of the teacher model vary for the same image in different iterations due to the random sampling in the LNM.

To address the above issues, we need to make the output of the teacher model become robust to the randomly sampled parameters $W$ and $Z$. Specifically, we propose an effective dual-path consistency training strategy. In teacher training, we duplicate the LNM module and the teacher model, with each path taking a noised label generated from independently sampled parameters. We then introduce a consistency loss between the outputs from these two paths. As a result, the final loss for the teacher training can be expressed as:
\begin{equation}
\begin{aligned}
\label{con:SL}
\mathcal{L}_T = \mathcal{L}_{seg}(O^1, Y) + \mathcal{L}_{seg}(O^2, Y) + \lambda \mathcal{D}(O^1, O^2),
\end{aligned}
\end{equation}
where the above $\mathcal{L}_{seg}(\cdot,\cdot)$ is the loss function for the segmentation task, $\lambda$ is loss coefficients to balance the contribution of the consistency loss and the task loss.

\subsection{Objective Function}

The training of the student model is the same as the standard knowledge distillation methods, where the model is supervised by both the label supervision and also supervision from the teacher model, which can be expressed as:
\begin{equation}
\begin{aligned}
\label{con:SL}
 \centering
 \mathcal{L}_{S} = \mathcal{L}_{seg}(O^S, Y) + \beta \mathcal{L}_{kd}(O^S, O^T) \; ,
\end{aligned}
\end{equation}
where $ O^S$ is the output of the student model, and $ O^T$ is the output of the teacher model, $\mathcal{L}_{kd}(\cdot,\cdot)$ can be an arbitrary distillation loss function. $\beta$ is the loss coefficient to balance the contribution of the distillation loss and the task loss. We only adopt the knowledge distillation for the logits map for simplicity.
% \begin{equation}
% \label{con:SL}
% \centering
%  L_{distill} = dist(pre_s, pre_t) 
% \end{equation}

% deep18  780  1018  415  1295  846 270
% psp18   758  944  402 1210 757 259

\begin{table*}[t]
    \scriptsize
    \centering
    \caption{Quantitative comparison of other distillation methods and our proposed approach, The results are reported in terms of mIoU on Cityscapes val dataset. All models are pretrained on ImageNet. $\mathrm{D_{time}}$ denotes the distillation time per iteration. $*$ denotes the model takes the noised label as privileged information.}
    \setlength{\tabcolsep}{.15mm}{
    \begin{tabular}{c|c|c|c|cc}
    %\midrule
    \hline
        \multirow{2}{*}{Method} & \multirow{2}{*}{\makecell[c]{Teacher (P (M) / \\ F (G) / val mIoU)}} & \multirow{2}{*}{\makecell[c]{Student (P (M) / \\ F (G) / (val/test mIoU) }} & \multirow{2}{*}{ \makecell[c]{$D_{\mathrm{time}}$ $\downarrow$ \\ (ms) }} & \multicolumn{2}{c}{mIoU $\uparrow$ } \\ 
         & & & & val & test \\

        %\midrule
        %\hline
        \hline
        \makecell[c]{SKD (TPAMI 2020)} & \multirow{5}[0]{*}{\makecell[c]{ DeepLabV3-R101 \\ (61.1\//2371.7\//78.1)} } & \multirow{5}[0]{*}{ \makecell[c]{DeepLabV3-R18 \\ (13.6\//572.0\//74.2/73.5)} } & 780 & 75.4(+1.2) & 74.1(+0.6)\\ 
        \makecell[c]{IFVD (ECCV 2020)} &  & & 1018 & 75.6(+1.4) & 74.3(+0.8)\\ 
        \makecell[c]{CWD (ICCV 2021)} &  & & 415 & 75.6(+1.4) & 74.1(+0.6)\\ 
        \makecell[c]{CIRKD (CVPR 2022)} &  & & 1295 & 76.4(+2.2) & 75.1(+1.6)\\ 
        \makecell[c]{MasKD (ICLR 2023)} &  & & 846 & 77.0(+2.8) & 75.6(+2.1)\\ 

        \hline
        \rowcolor{gray!15} Ours & \colorbox{gray!15}{\makecell[c]{DeepLabV3-R18* \\ (13.6\//572.1\//79.8)}} & \colorbox{gray!15}{\makecell[c]{DeepLabV3-R18 \\ (13.6\//572.0\//74.2/73.5)}} & \textbf{270} & \textbf{77.7(+3.5)} & \textbf{76.9(+3.4)} \\ 
        %\midrule
        \hline
        \hline
        \makecell[c]{SKD (TPAMI 2020)} & \multirow{5}[0]{*}{\makecell[c]{ DeepLabV3-R101 \\ (61.1\//2371.7\//78.1)} } & \multirow{5}[0]{*}{ \makecell[c]{PSPNet-R18 \\ (12.9\//507.4\//72.6/72.3)} } & 758 & 73.3(+0.7) & 73.0(+0.7) \\ 
        \makecell[c]{IFVD (ECCV 2020)} &  & & 944 & 73.7(+1.1) & 72.8(+0.5) \\ 
        \makecell[c]{CWD (ICCV 2021)} &  &  & 402 & 74.4(+1.8) & 73.6(+1.3)\\ 
        \makecell[c]{CIRKD (CVPR 2022)} &  & & 1210 & 74.7(+2.1) & 74.1(+1.8) \\ 
        \makecell[c]{MasKD (ICLR 2023)} &  & & 757 & 75.3(+2.7) & 74.6(+2.3)\\ 
        
        \hline
        \rowcolor{gray!15} Ours & \colorbox{gray!15}{\makecell[c]{PSPNet-R18*  \\ (12.9\//507.5\//79.7)}} & \colorbox{gray!15}{\makecell[c]{PSPNet-R18 \\ (12.9\//507.4\//72.6/72.3)}} & \textbf{259} & \textbf{76.2(+3.6)} & \textbf{75.6(+3.3)} \\ 
        \hline
    \end{tabular}}
    \label{tab:maincomp}%
\end{table*}

\section{Experiments}

\subsection{Experimental Setup}

For thorough evaluation, we conduct experiments on five commonly used datasets: Cityscape \cite{cordts2016cityscapes}, ADE20K \cite{zhou2017scene}, and PASCAL-VOC \cite{everingham2010pascal}, COCO-Stuff 10K \cite{caesar2018coco} and COCO-Stuff 164K \cite{caesar2018coco} datasets, and five different popular segmentation baseline models, including FCN \cite{long2015fully}, PSPNet \cite{zhao2017pyramid}, DeeplabV3 \cite{chen2017rethinking}, STDC \cite{fan2021rethinking}, and OCRNet \cite{yuan2020object}. We use the commonly used mean Intersection over Union (mIoU) metric for evaluation. We adopt the CWD distillation loss \cite{shu2021channel} as $\mathcal{L}_{kd}(\cdot,\cdot)$ in our approach as it is a convenient plug-in loss and also very efficient and effective. For simplify, we also use the CWD loss as the consistent loss in the teacher model training. We set the hyper-parameters $\alpha, \lambda, \beta$ as 0.01, 1, and 3, respectively. More details about training setting can be found in the supplementary material.

\begin{table}[t]
    \scriptsize
    \centering

  \floatsetup{floatrowsep=qquad, captionskip=4pt}
  \begin{floatrow}[2]
    \ttabbox%
    {\setlength{\tabcolsep}{.0mm}{\begin{tabularx}{0.48\textwidth}{c|c|c|c}
    %\midrule
    \hline
        \multirow{2}{*}{Method} & \multirow{2}{*}{ \makecell[c]{Teacher \\ mIoU (\%)}} & \multirow{2}{*}{ \makecell[c]{Student \\ mIoU  (\%)}} & \multirow{2}{*}{mIoU} \\ 
         & & &\\

        %\midrule 
        %\hline
        \hline
        \makecell[c]{SKD} & \multirow{4}[0]{*}{\makecell[c]{ DeepLabV3-\\R101 (78.1)} } & \multirow{4}[0]{*}{ \makecell[c]{DeepLabV3\\-R18 (61.8)} }  & 62.7(+0.9) \\ 
        \makecell[c]{IFVD} &  &  & 63.1(+1.3) \\ 
        \makecell[c]{CWD} &  &  & 64.6(+2.8) \\ 
        \makecell[c]{CIRKD} &  &  & 67.4(+5.6) \\ 
        
        \hline
        \rowcolor{gray!15} Ours & \colorbox{gray!15}{\makecell[c]{DeepLabV3-\\R18* (79.8)}} & \colorbox{gray!15}{\makecell[c]{DeepLabV3\\-R18  (61.8) }} &  \textbf{68.1(+6.3)}  \\ 
        \midrule

        \hline
        \makecell[c]{SKD} & \multirow{4}[0]{*}{\makecell[c]{ DeepLabV3-\\R101 (78.1)} } & \multirow{4}[0]{*}{ \makecell[c]{PSPNet-\\R18 (65.2)} }  & 67.1(+1.9) \\ 
        \makecell[c]{IFVD} &  &  & 66.0(+0.8) \\ 
        \makecell[c]{CWD} &  &  & 67.7(+2.5) \\ 
        \makecell[c]{CIRKD} &  &  & 68.2(+3.0) \\ 
      
        \hline
        \rowcolor{gray!15} Ours & \colorbox{gray!15}{\makecell[c]{PSPNet-\\R18* (79.7) }} & \colorbox{gray!15}{\makecell[c]{PSPNet-\\R18 (65.2) }} & \textbf{69.2(+4.0)} \\ 
        \hline
      \end{tabularx}}}
    {\caption[Valori medi]{Quantitative comparison  on Cityscapes val set. Student models are trained from scratch. $*$ denotes model takes noised label as privileged information.}
      \label{tab:compscratch}}
    %\hfill%
    \hspace{-0.3cm}
    \ttabbox%
    {\setlength{\tabcolsep}{.0mm}{\begin{tabularx}{0.48\textwidth}{c|c|c|c}
    %\midrule
    \hline
        \multirow{2}{*}{Method} & \multirow{2}{*}{ \makecell[c]{Teacher \\ mIoU (\%)}} & \multirow{2}{*}{ \makecell[c]{Student \\ mIoU  (\%)}} & \multirow{2}{*}{mIoU} \\ 
         & & &\\

        %\midrule
        %\hline
        \hline
        \makecell[c]{SKD} & \multirow{4}[0]{*}{\makecell[c]{ DeepLabV3-\\R101 (77.7)} } & \multirow{4}[0]{*}{ \makecell[c]{DeepLabV3\\-R18 (73.2)} }  & 73.5(+0.2) \\ 
        \makecell[c]{IFVD} &  &  & 73.9(+0.7) \\ 
        \makecell[c]{CWD} &  &  & 74.0(+0.8) \\ 
        \makecell[c]{CIRKD} &  &  & 74.5(+1.3) \\ 
       
        \hline
        \rowcolor{gray!15} Ours & \colorbox{gray!15}{\makecell[c]{DeepLabV3-\\R18* (79.1)}} & \colorbox{gray!15}{\makecell[c]{DeepLabV3\\-R18  (73.2) }} &  \textbf{75.0(+1.8)}  \\ 
        \midrule

        \hline
        \makecell[c]{SKD} & \multirow{4}[0]{*}{\makecell[c]{ DeepLabV3-\\R101 (77.7)} } & \multirow{4}[0]{*}{ \makecell[c]{PSPNet-\\R18 (73.3)} }  & 74.1(+0.8) \\ 
        \makecell[c]{IFVD} &  &  & 73.5(+0.3) \\ 
        \makecell[c]{CWD} &  &  & 74.0(+0.7) \\ 
        \makecell[c]{CIRKD} &  &  & 74.8(+1.5) \\ 
       
        \hline
        \rowcolor{gray!15} Ours & \colorbox{gray!15}{\makecell[c]{PSPNet-\\R18* (79.1) }} & \colorbox{gray!15}{\makecell[c]{PSPNet-\\R18 (73.3) }} & \textbf{75.4(+2.1)}  \\ 
        
        \hline
      \end{tabularx}}}
    {\caption[Valori medi]{Quantitative comparison on the PASCAL VOC dataset. $*$ denotes the teacher model takes the noised label as privileged information.}
      \label{tab:comppascalvoc}}
  \end{floatrow}
\end{table}%

\subsection{Main Results}
\label{ana:mainab}
\textbf{Comparison with SOTA Methods.}
We compare our proposed distillation approach with several state-of-the-art methods, including SKD\cite{liu2020structured}, IFVD\cite{wang2020intra}, CWD\cite{shu2021channel}, CIRKD\cite{yang2022cross}, MasKD\cite{huang2022masked}. They all rely on more complex models to provide additional knowledge in distillation training. The experimental results are shown in Tab. \ref{tab:maincomp}. It can be seen that for the PSPNet-R18 model, we achieve very competitive results among all other distillation approaches, we suppress the CWD baseline approach by 1.8\% mIoU on the val set and 2.0\% mIoU on the test set,  For the DeepLabV3-R18 model, it can be seen that we achieve the best performance among all the approaches and suppress the CWD baseline approach by 2.1\% mIoU and 2.8\% mIoU, respectively. Additionally, we conducted distillation time comparison on NVIDIA V100 GPU with image size of $512 \times 1024$, it can be seen that our method shows a significant advantage over others.

Tab. \ref{tab:compscratch} shows the results of the performance comparison when the student model is trained from scratch. It can be seen that our approach achieves the best results among all other methods. For the PSPNet-R18 model, we suppress the baseline CWD by 1.5\% mIoU, and suppress the CIRKD method by 1.0\% mIoU. For the DeepLabV3-R18 model, we suppress the baseline CWD by 3.5\% mIoU, and suppress the CIRKD method by 0.7\% mIoU. Tab. \ref{tab:comppascalvoc} shows the experimental results on the PASCAL-VOC datasets, we can see that our approach also shows better performance compared with other methods.

We also conduct experiments on other six models on the Cityscapes dataset, results are shown in Tab. \ref{tab:moremodel}. For FCN-18, we suppress the baseline by 1.6\%, for FCN-50,  we suppress the baseline by 0.7\%. For the STDC baseline, we suppress the STDC1 and STDC2 by 1.3\% and 1.6\% mIoU. For the OCRNet baseline,  the experiments are conducted on models with OCRNet-W18s and OCRNet-W18 as the backbone, we suppress the baseline by 1.2\% and 1.8\% mIoU, respectively.

To further prove the generalization of our approach, we conduct experiments on ADE20K \cite{zhou2017scene}, PASCAL-VOC \cite{everingham2010pascal},  COCO-Stuff 10K \cite{caesar2018coco} and COCO-Stuff 164K \cite{caesar2018coco} datasets using PSPNet and DeepLabV3 methods with ResNet50 backbone. The results are shown in Tab. \ref{tab:apexp}. It is evident that our method significantly enhances the performance of baseline models across various datasets.

\begin{table}[t]
\scriptsize
  \centering
  \caption{Quantitative comparison between CWD baseline and our approach on more segmentation models. Our teacher model is the same model as the student model but trained with the noised label as input.}
  \setlength{\tabcolsep}{1.5mm}{
    \begin{tabular}{l|cccc|ccc}
    \hline
    Teacher & \multicolumn{4}{c|}{DeepLabV3-R101} & \multicolumn{2}{c}{OCRNet-W48} \\
    \hline
    Student & \makecell[c]{FCN-18 } & \makecell[c]{FCN-50 } & \makecell[c]{STDC1 } & \makecell[c]{STDC2 } & \makecell[c]{OCR-W18 } & \makecell[c]{OCR-W18 } \\
    \hline
    None  & 69.7  & 73.8  & 71.8  & 73.2 & 74.3  & 77.7 \\
    %\hline
    CWD (base)  & 70.8  & 74.4  & 72.3  & 74.1 & 76.9  & 78.6 \\
    %\hline
    \rowcolor{gray!15}Ours   & \textbf{71.3}  & \textbf{74.5}  & \textbf{73.1}  & \textbf{74.8} & \textbf{78.1}  & \textbf{80.4}  \\
    \hline
    \end{tabular}%
    }
  \label{tab:moremodel}%
\end{table}%

\begin{table*}[t]
\scriptsize
\centering
\caption{Quantitative results on more datasets. Our teacher model is the same model as the student model but trained with the noised label as input. PSP50 is the abbreviation for PSPNet-R50, DeepV3-50 is the abbreviation for DeepLabV3-R50.}
\setlength{\tabcolsep}{.75mm}{
\begin{tabular}{l|cc|cc|cc|cc}
%\midrule
\hline
Dataset & \multicolumn{2}{c|}{ADE20K} & \multicolumn{2}{c|}{PASCAL-VOC 2012} & \multicolumn{2}{c|}{COCO-Stuff 10K} & \multicolumn{2}{c}{COCO-Stuff 164K} \\
\hline
Student & \makecell[c]{ PSP50 } & \makecell[c]{DeepV3-50 } & \makecell[c]{ PSP50 } & \makecell[c]{DeepV3-50 } & \makecell[c]{ PSP50 } & \makecell[c]{DeepV3-50 } & \makecell[c]{ PSP50 } & \makecell[c]{DeepV3-50 }  \\
%\midrule
%\midrule
\hline
None & 41.1  & 42.4 & 76.8  & 76.3 & 35.7  & 34.7 & 38.8  & 39.4 \\

CWD (base) & 41.3  &  42.2 &  77.2 &  76.8 & 35.4  &  35.5 &  39.2 &  39.4 \\

\rowcolor{gray!15} Ours  & \textbf{42.0}  & \textbf{43.9} & \textbf{77.6}  & \textbf{77.5} & \textbf{36.0}  & \textbf{35.6} & \textbf{39.9}  & \textbf{40.1}  \\
%\midrule
\hline
\end{tabular}%
}
\label{tab:apexp}%
\end{table*}

\noindent \textbf{Comparison with Stronger Teacher.}  As can be seen from Tab. \ref{tab:maincomp}, after introducing noised label to the input, we obtain lightweight teacher models with very strong performance, even surpassing DeepLabV3-R101. However, this raises a question about the underlying reasons for our method's effectiveness: Is the performance boost due to a more powerful teacher model? To answer this question, we chose other two powerful models as a teacher: PSPNet-R101 (79.8 mIoU) and OCRNet-W48 (80.7 mIoU). The experimental results are shown in Tab. \ref{tab:wit}. It can be seen that distillation with a more powerful teacher model does bring some performance improvements, but our method still retains a clear advantage over them. Here we give an intuitive explanation about why our performance is better than distilling with stronger teachers: the knowledge that comes from the same model structure is more transferable between each other. For more detailed experiments and discussion, please refer to the supplementary material.

\begin{table}[t]
\centering
\scriptsize
\setlength{\extrarowheight}{1.6pt} 
\setlength{\tabcolsep}{.4mm}{
\begin{tabular}{c|c|c|c|c|c|c}
\hline
    \multicolumn{1}{c|}{\multirow{2}{*}{\makecell[c]{Teacher \\ (mIoU) }}} & \multirow{2}{*}{\makecell[c]{None}} & \multicolumn{1}{c|}{ \multirow{2}{*}{\makecell[c]{DeepV3-R101 \\ (78.1)}} } & \multicolumn{1}{c|}{ \multirow{2}{*}{\makecell[c]{PSPNet-R101 \\ (79.8)}} } &  \multicolumn{1}{c|}{ \multirow{2}{*}{\makecell[c]{OCRNet-W48 \\ (80.7)}}  }  & \multirow{2}{*}{\makecell[c]{PSPNet-R18$*$\\ (79.7)}}  & \multirow{2}{*}{\makecell[c]{DeepV3-R18$*$\\ (79.8)}}  \\ 
    
    & & & & & & \\ \hline
    
    PSPNet-R18 & 72.6 & 74.4(+1.8) & 75.9(+3.3) & 75.0(+2.4) & \textbf{76.2(+3.6)} & \textbf{76.2(+3.6)}  \\
    %\cline {2 - 7}
    DeepLabV3-R18 & 74.2 & 75.6(+1.4) & 77.1(+2.9) & 77.2(+3.0) & 77.4(+3.2) & \textbf{77.7(+3.5)}  \\
    \hline
\end{tabular}}
 \caption{Experimental analysis about the reason of performance gains of the student models. The experiments are conducted on the Cityscapes val set. $*$ denotes the model takes the noised label as privileged information.}
\label{tab:wit}
\end{table}

\noindent \textbf{Simple to Complex Model Distillation.} As the performance of a model can be significantly improved with the privileged information, the performance of a simple model with privileged information can be even higher than a complex model. As shown in Tab. \ref{tab:maincomp}, the performance of the PSPNet-R18 with label information as input is 79.7\%, surpassing that of the DeepLabV3 model with a ResNet101 backbone. So we wonder if we can distill knowledge from a label-assisted simple teacher to a complex student. To explore this, we conducted experiments using the STDC models, with the results presented in Tab. \ref{tab:sccs}. Here we focus on the result in the third row, we can see that with the simple STDC1 model as a teacher and the complex STDC2 as a student, the performance of STDC2 model still can be improved by 1.4 \% mIoU. The experimental results confirm that complex models can indeed benefit from learning from label-assisted simpler teacher models. 

As large foundational models in vision gain popularity, we also extended our experiments to assess the applicability of our method to these models. We employ a relatively lightweight yet high-performance model (OCRNet-W48) with noised label input as teacher and used LoRA \cite{hu2021lora} method to finetune DINOv2 \cite{oquab2023dinov2} on VOC dataset. The results also show performance improvements:
 81.1 $\rightarrow$ \textbf{82.3} (with backbone ViT-S/14) and
 82.5 $\rightarrow$ \textbf{83.2} (with backbone ViT-B/14),
 indicating that the teacher model with noised labels as input contains knowledge that is not present in large models. 
 
In addition, as the teacher can be the same as the student, so our approach can also be seen as a self-enhanced technique, which provides greater potential in the future. Please refer to the supplementary material for more information.

\begin{table}[t]
    \scriptsize
    \centering

  \floatsetup{floatrowsep=qquad, captionskip=4pt}
  \begin{floatrow}[2]
    \ttabbox%
    {\begin{tabularx}{0.478\textwidth}{c|cc|c}
    %\midrule
    \hline
    Mode & Teacher   & Student  & mIoU  \\
    %\midrule
    \hline
    \multirow{2}[0]{*}{\makecell[c]{simple->simple\\complex->complex}} & M1 (79.1) & M1 (71.8) & 73.1 \\
         & M2 (80.0) & M2 (73.2) & 74.8 \\
    %\midrule
    \hline
    \multirow{2}[0]{*}{\makecell[c]{simple->complex \\ complex->simple}} & M1 (79.1) & M2 (73.2) & 74.6 \\
      & M2 (80.0) & M1 (71.8) & 73.5 \\
    %\midrule
    \hline
      \end{tabularx}}
    {\caption[Valori medi]{Experimental of distillation between models of different complexities. M is short for STDC model. The results are reported on Cityscapes val dataset.}
      \label{tab:sccs}}
    \hfill%
    \hspace{-0.3cm}
    \ttabbox%
    {\begin{tabularx}{0.479\textwidth}{c|c|c|c}
    %\midrule
    %\midrule
    \hline
    Model & $\mathcal{D}(\cdot,\cdot)$  &  mKL ($\times$ 100) & mIoU \\
    
    %\midrule
    \hline
    \makecell[c]{DeepLabV3-R18} &  & 6.8 & 76.7 \\
    \makecell[c]{DeepLabV3-R18} & \checkmark & 3.4 & 77.7\\
    %\midrule
    \hline
    \makecell[c]{STDC2} &   & 3.4 & 73.6\\
    \makecell[c]{STDC2} & \checkmark & 2.1 & 74.8 \\
    %\midrule
    \hline
      \end{tabularx}}
    {\caption[Valori medi]{Ablation study on consistency loss for teacher. $\times$100 for better presentation. mIoU is student performance when distillation with the corresponding teacher.}
    \label{tab:consisab}}
  \end{floatrow}
\end{table}%

\begin{figure*}[t]
\centering
\includegraphics[width=1.\linewidth]{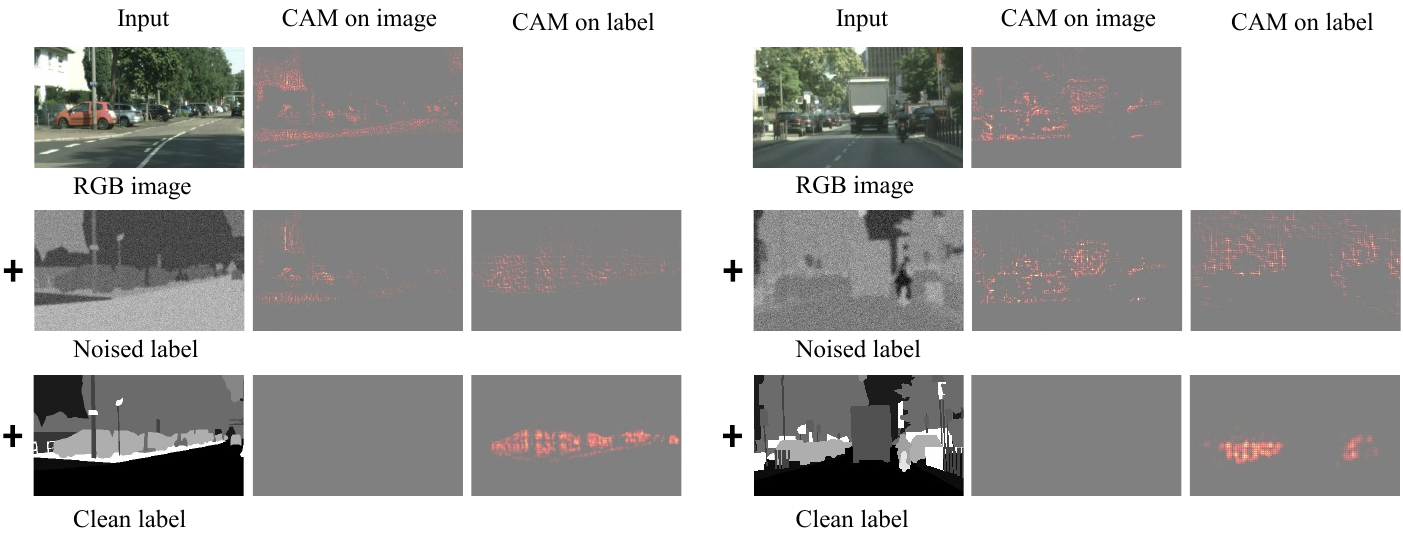}
\caption{Experiment on the contribution of different inputs to the predictions. Here we choose the contribution to the class \emph{car} for visualization, where red regions correspond to high contribution for the class. We use LayerCAM \cite{jiang2021layercam} for its better performance in lower layers. Best viewed with zoomed in.}
\label{fig1:shortcut}
\end{figure*}

\noindent \textbf{Shortcut Learning with Clean Label Input.} To demonstrate that the model will learn a shortcut instead of extracting useful information when we take clean label as input in training, we conduct experiments to assess the contribution of two different inputs to the final prediction. The results are shown in Fig. \ref{fig1:shortcut}. It can be seen that when we input with noised label, the contribution of the final prediction obviously comes from two parts. However, when we input with clean label, the prediction predominantly relies on the input label alone, with the RGB image contributing negligibly to the output. Therefore, in this case, the model has learned a mapping from input label to output label, rather than extracting useful information from the RGB image as we have mentioned in subsection \ref{sec:probformu}.

\subsection{Ablation Studies}

\textbf{Dual-path Consistency Loss.} We perform ablation studies to verify the effectiveness of the dual-path consistency training strategy. We conduct the experiments on models DeeplabV3-18 and STDC2. To assess the robustness of final the teacher model against the label noising operation, we perform random sampling $m$ times for the transformation parameters and obtain the corresponding output logits of the teacher with the same image $X_i$, referred as $O_i^1, O_i^2,..., O_i^m$, where $O_i^k \in \mathbb{R}^{C \times H \times W}$. We then calculate the $KL$ divergence of distributions between each pair. We utilize the mean value of the distance as a criterion of stability: $KL_{mean}=\frac{1}{m\cdot m}\sum_{i=k}^{m} \sum_{j=1}^{m} KL(softmax(O_i^k,dim\hspace{-0.1cm}=\hspace{-0.1cm}0), softmax(O_i^j,dim\hspace{-0.1cm}=\hspace{-0.1cm}0)) $. In our experiment, we set $n=3$. The results are shown in Tab. \ref{tab:consisab}. show that our consistent loss significantly reduces the distribution distance between outputs, enhancing the stability of the distillation process and notably improving the performance of the student model.

% It can be seen that with our consistent loss, the distribution distance between the outputs can be reduced significantly so that the distillation process will be more stable. Thus the performance of student model will also have a more obvious improvement.

\begin{table}[t]
%\footnotesize
\scriptsize
  \centering
  \caption{Sensitive analysis about the class-wise and pixel-wise label nosing operation. The experiments are conducted with the STDC2 model on the cityscapes val set.}
  \setlength{\tabcolsep}{1.5mm}{
    \begin{tabular}{c|ccccc|ccccc}
    %\midrule
    \hline
    & \multicolumn{5}{c|}{ \makecell[c]{with class-wise  noising}} & \multicolumn{4}{c}{\makecell[c]{w/o class-wise noising}} \\
    %\midrule
    \hline
    \makecell[c]{Pixel-wise noising ($\alpha$)} & 0 & 1e-3 & 1e-2  & 1e-1 & 1.0 & 0 &1e-3 & 1e-2  & 1e-1   & 1.0    \\
    %\midrule
    \hline
    Teacher (mIoU) & 80.5 & 80.5  & 80.0  & 74.1 &  71.7 & 90.0 & 84.3  & 86.5  & 83.2  & 76.3   \\
    \rowcolor{gray!15} Student (mIoU) & 73.8 & 74.5  & 74.8  & 72.3 &  70.5 & 69.4  & 70.1  & 72.3  & 73.0    & 68.8   \\
    %\midrule
    \hline
    \end{tabular}%
    }
  \label{tab:cwpwab}%
\end{table}%

\noindent \textbf{Sensitivity Analysis.}
We conducted experiments to evaluate the effects of class-wise and pixel-wise noising on the performance of both the teacher and student models. The results are shown in Tab. \ref{tab:cwpwab}. It should be noted that if without class-wise noising operation, the labels are first normalized to achieve a similar input interval with the normalized images. With class-wise noising, if the pixel-wise noise is lower (0.001), better performance can be achieved for the teacher model as the information is cleaner. However, this knowledge cannot be effectively transferred as the predictions rely more on the leaked label information. When the pixel-wise noise is higher (0.1), the useful information in the noised label will be contaminated by the noise, degrading the teacher model’s performance and subsequently affecting the performance of the student model. Without class-wise noising operation, the teacher model can achieve higher performance as the class index information is preserved. However, as predictions are more dependent on label information than on extracted features, the knowledge in the outputs cannot be effectively transferred to the student model, resulting in unsatisfactory distillation performance. 

\subsection{Discussion}
\label{sec:edp}
Here, we offer an alternative denoising perspective to explain our approach of utilizing noised label as privileged information. For the teacher model, it receives both the image $X_i$ and the noised label $\phi(Y_i|\theta)$ as input, aiming to produce the clean label $Y_i$ as its output. Hence, the teacher model can be considered  as a denoising model whose objective is to eliminate the noise present in the input noised label. In this case, the objective of model learning is changed from
%This changes the explanation of the forward process from:
\begin{equation}
\begin{aligned}
\label{con:SL}
\centering
{Y_i} = mapping(X_i),
\end{aligned}
\end{equation}  
to
\begin{equation}
\begin{aligned}
\label{con:SL}
\centering
{Y_i} = denoising(X_i, \phi(Y_i|\theta)),
\end{aligned}
\end{equation}  
where the behavior of noising function $\phi(\cdot)$ is controlled by the parameter $\theta$, which is randomly sampled every iteration in training. The above equation shows that we transform the standard segmentation task of learning an image-to-label mapping to a label-denoising task with the RGB images serve as observations.

From the above perspective, the performance of the denoising task should heavily depend on the level of noise in the input, which aligns with our experiment results shown in Tab. \ref{tab:cwpwab}. The results indicated that performance varied significantly depending on the noise levels. With relatively minimal noise (0.001 for pixel-wise noising), the teacher model could achieve high performance effortlessly. However, when noise levels were substantially increased (0.1 for pixel-wise noising), the performance of the teacher model declined dramatically. This decline is primarily due to the model's difficulty in extracting useful information from the excessively noised labels.

\section{Conclusion}
In conclusion, our proposed knowledge distillation approach for semantic segmentation offers a novel and effective way to leverage readily available labels as a source of knowledge. By strategically introducing noise to labels and employing a dual-path consistency training strategy, we significantly enhance the performance of lightweight teacher models. Extensive experiments demonstrate the effectiveness of our method, achieving substantial improvements in mIoU scores across various datasets and baseline models. We believe this work provides valuable insights and a promising direction for future research in knowledge distillation for semantic segmentation.
%In this paper, we present an efficient and effective knowledge distillation technique tailored for semantic segmentation tasks. Different from the previous approaches that rely on computationally costly power-pretrained teacher models or other modalities, our approach leverages labels, which is readily available during training as a source of additional knowledge. Specifically, we propose to noise the clean label and then use them as privileged information to boost the performance of the lightweight teacher model. To improve the robustness of the teacher model against noise in the privileged information. we further propose a dual-path consistency training strategy with distance loss between the outputs of two paths. We maintain the student model training the same as the standard framework for simplicity. Finally, we conduct experiments on five challenging datasets and five different baseline models, a substantial and consistent improvement is achieved in the mIoU score. We also perform analysis on the crucial components of our approach and hope that other researchers can get inspiration from our study. 

% Our future work is to extend our approach to other 2D/3D tasks, such as panoptic segmentation, 3D point cloud semantic segmentation, and object detection.

\section{Limitations}
We believe that our proposed knowledge distillation approach can significantly enhance the performance of baseline models across various computer vision tasks.  Our approach has potential value in the deployment of deep learning models in practice, we do not see any potential negative effects.

\section*
{Acknowledgements}

This work is supported by the Shanghai Platform for Neuromorphic and AI Chip under Grant 17DZ2260900 (NeuHelium). The computations in this research were performed using the CFFF platform of Fudan University.

\bibliographystyle{splncs04}
\bibliography{main}

\begin{thebibliography}{10}
\providecommand{\url}[1]{\texttt{#1}}
\providecommand{\urlprefix}{URL }
\providecommand{\doi}[1]{https://doi.org/#1}

\bibitem{alzahrani2021biomedical}
Alzahrani, Y., Boufama, B.: Biomedical image segmentation: a survey. SN Computer Science  \textbf{2},  1--22 (2021)

\bibitem{caesar2018coco}
Caesar, H., Uijlings, J., Ferrari, V.: Coco-stuff: Thing and stuff classes in context. In: CVPR. pp. 1209--1218 (2018)

\bibitem{chen2021cross}
Chen, D., Mei, J.P., Zhang, Y., Wang, C., Wang, Z., Feng, Y., Chen, C.: Cross-layer distillation with semantic calibration. In: AAAI. vol.~35, pp. 7028--7036 (2021)

\bibitem{chen2017rethinking}
Chen, L.C., Papandreou, G., Schroff, F., Adam, H.: Rethinking atrous convolution for semantic image segmentation. arXiv preprint arXiv:1706.05587  (2017)

\bibitem{cordts2016cityscapes}
Cordts, M., Omran, M., Ramos, S., Rehfeld, T., Enzweiler, M., Benenson, R., Franke, U., Roth, S., Schiele, B.: The cityscapes dataset for semantic urban scene understanding. In: CVPR. pp. 3213--3223 (2016)

\bibitem{das2023understanding}
Das, R., Sanghavi, S.: Understanding self-distillation in the presence of label noise. arXiv preprint arXiv:2301.13304  (2023)

\bibitem{everingham2010pascal}
Everingham, M., Van~Gool, L., Williams, C.K., Winn, J., Zisserman, A.: The pascal visual object classes (voc) challenge. International journal of computer vision  \textbf{88},  303--338 (2010)

\bibitem{fan2021rethinking}
Fan, M., Lai, S., Huang, J., Wei, X., Chai, Z., Luo, J., Wei, X.: Rethinking bisenet for real-time semantic segmentation. In: CVPR. pp. 9716--9725 (2021)

\bibitem{feyereisl2012privileged}
Feyereisl, J., Aickelin, U.: Privileged information for data clustering. Information Sciences  \textbf{194},  4--23 (2012)

\bibitem{geirhos2020shortcut}
Geirhos, R., Jacobsen, J.H., Michaelis, C., Zemel, R., Brendel, W., Bethge, M., Wichmann, F.A.: Shortcut learning in deep neural networks. Nature Machine Intelligence  \textbf{2}(11),  665--673 (2020)

\bibitem{he2019adaptive}
He, J., Deng, Z., Zhou, L., Wang, Y., Qiao, Y.: Adaptive pyramid context network for semantic segmentation. In: CVPR. pp. 7519--7528 (2019)

\bibitem{he2019knowledge}
He, T., Shen, C., Tian, Z., Gong, D., Sun, C., Yan, Y.: Knowledge adaptation for efficient semantic segmentation. In: CVPR. pp. 578--587 (2019)

\bibitem{heo2022vita}
Heo, M., Hwang, S., Oh, S.W., Lee, J.Y., Kim, S.J.: Vita: Video instance segmentation via object token association. NIPS  \textbf{35},  23109--23120 (2022)

\bibitem{hinton2015distilling}
Hinton, G., Vinyals, O., Dean, J.: Distilling the knowledge in a neural network. arXiv preprint arXiv:1503.02531  (2015)

\bibitem{hu2023teacher}
Hu, C., et~al.: Teacher-student architecture for knowledge distillation: A survey. arXiv:2308.04268  (2023)

\bibitem{hu2021lora}
Hu, E.J., Shen, Y., Wallis, P., Allen-Zhu, Z., Li, Y., Wang, S., Wang, L., Chen, W.: Lora: Low-rank adaptation of large language models. arXiv preprint arXiv:2106.09685  (2021)

\bibitem{hu2020knowledge}
Hu, M., Maillard, M., Zhang, Y., Ciceri, T., La~Barbera, G., Bloch, I., Gori, P.: Knowledge distillation from multi-modal to mono-modal segmentation networks. In: MICCAI. pp. 772--781. Springer (2020)

\bibitem{huang2022masked}
Huang, T., Zhang, Y., You, S., Wang, F., Qian, C., Cao, J., Xu, C.: Masked distillation with receptive tokens. In: ICLR (2023)

\bibitem{huang2022label}
Huang, Y., Liu, X., Zhu, Y., Xu, Z., Shen, C., Che, Z., Zhang, G., Peng, Y., Feng, F., Tang, J.: Label-guided auxiliary training improves 3d object detector. In: ECCV. pp. 684--700. Springer (2022)

\bibitem{jiang2021layercam}
Jiang, P.T., Zhang, C.B., Hou, Q., Cheng, M.M., Wei, Y.: Layercam: Exploring hierarchical class activation maps for localization. IEEE Transactions on Image Processing  \textbf{30},  5875--5888 (2021)

\bibitem{lee2020learning}
Lee, W., Lee, J., Kim, D., Ham, B.: Learning with privileged information for efficient image super-resolution. In: ECCV. pp. 465--482. Springer (2020)

\bibitem{li2016large}
Li, A., Lu, Z., Wang, L., Han, P., Wen, J.R.: Large-scale sparse learning from noisy tags for semantic segmentation. IEEE transactions on cybernetics  \textbf{48}(1),  253--263 (2016)

\bibitem{li2019dfanet}
Li, H., Xiong, P., Fan, H., Sun, J.: Dfanet: Deep feature aggregation for real-time semantic segmentation. In: CVPR. pp. 9522--9531 (2019)

\bibitem{li2020enhanced}
Li, X., Li, W., Ren, D., Zhang, H., Wang, M., Zuo, W.: Enhanced blind face restoration with multi-exemplar images and adaptive spatial feature fusion. In: CVPR. pp. 2706--2715 (2020)

\bibitem{li2017learning}
Li, Y., Yang, J., Song, Y., Cao, L., Luo, J., Li, L.J.: Learning from noisy labels with distillation. In: ICCV. pp. 1910--1918 (2017)

\bibitem{liu2020structured}
Liu, Y., Shu, C., Wang, J., Shen, C.: Structured knowledge distillation for dense prediction. TPAMI  (2020)

\bibitem{liu20213d}
Liu, Z., Qi, X., Fu, C.W.: 3d-to-2d distillation for indoor scene parsing. In: CVPR. pp. 4464--4474 (2021)

\bibitem{long2015fully}
Long, J., Shelhamer, E., Darrell, T.: Fully convolutional networks for semantic segmentation. In: CVPR. pp. 3431--3440 (2015)

\bibitem{lopez2015unifying}
Lopez-Paz, D., Bottou, L., Sch{\"o}lkopf, B., Vapnik, V.: Unifying distillation and privileged information. arXiv preprint arXiv:1511.03643  (2015)

\bibitem{lu2016learning}
Lu, Z., Fu, Z., Xiang, T., Han, P., Wang, L., Gao, X.: Learning from weak and noisy labels for semantic segmentation. IEEE transactions on pattern analysis and machine intelligence  \textbf{39}(3),  486--500 (2016)

\bibitem{nguyen2021dataset}
Nguyen, T., Novak, R., Xiao, L., Lee, J.: Dataset distillation with infinitely wide convolutional networks. NIPS  \textbf{34},  5186--5198 (2021)

\bibitem{oquab2023dinov2}
Oquab, M., Darcet, T., Moutakanni, T., Vo, H., Szafraniec, M., Khalidov, V., Fernandez, P., Haziza, D., Massa, F., El-Nouby, A., et~al.: Dinov2: Learning robust visual features without supervision. arXiv preprint arXiv:2304.07193  (2023)

\bibitem{ronneberger2015u}
Ronneberger, O., Fischer, P., Brox, T.: U-net: Convolutional networks for biomedical image segmentation. In: MICCAI. pp. 234--241. Springer (2015)

\bibitem{scimeca2021shortcut}
Scimeca, L., Oh, S.J., Chun, S., Poli, M., Yun, S.: Which shortcut cues will dnns choose? a study from the parameter-space perspective. In: ICLR (2021)

\bibitem{shu2021channel}
Shu, C., Liu, Y., Gao, J., Yan, Z., Shen, C.: Channel-wise knowledge distillation for dense prediction. In: ICCV. pp. 5311--5320 (2021)

\bibitem{siam2018comparative}
Siam, M., Gamal, M., Abdel-Razek, M., Yogamani, S., Jagersand, M., Zhang, H.: A comparative study of real-time semantic segmentation for autonomous driving. In: CVPRW. pp. 587--597 (2018)

\bibitem{stanton2021does}
Stanton, S., Izmailov, P., Kirichenko, P., Alemi, A.A., Wilson, A.G.: Does knowledge distillation really work? NIPS  \textbf{34},  6906--6919 (2021)

\bibitem{strudel2021segmenter}
Strudel, R., Garcia, R., Laptev, I., Schmid, C.: Segmenter: Transformer for semantic segmentation. In: ICCV. pp. 7262--7272 (2021)

\bibitem{vapnik2009new}
Vapnik, V., Vashist, A.: A new learning paradigm: Learning using privileged information. Neural networks  \textbf{22}(5-6),  544--557 (2009)

\bibitem{vobecky2022drive}
Vobecky, A., Hurych, D., Sim{\'e}oni, O., Gidaris, S., Bursuc, A., P{\'e}rez, P., Sivic, J.: Drive\&segment: Unsupervised semantic segmentation of urban scenes via cross-modal distillation. In: ECCV. pp. 478--495. Springer (2022)

\bibitem{wang2022rtformer}
Wang, J., Gou, C., Wu, Q., Feng, H., Han, J., Ding, E., Wang, J.: Rtformer: Efficient design for real-time semantic segmentation with transformer. NIPS  \textbf{35},  7423--7436 (2022)

\bibitem{wang2021knowledge}
Wang, L., Yoon, K.J.: Knowledge distillation and student-teacher learning for visual intelligence: A review and new outlooks. IEEE transactions on pattern analysis and machine intelligence  (2021)

\bibitem{wang2020intra}
Wang, Y., Zhou, W., Jiang, T., Bai, X., Xu, Y.: Intra-class feature variation distillation for semantic segmentation. In: ECCV. pp. 346--362. Springer (2020)

\bibitem{xiao2018unified}
Xiao, T., Liu, Y., Zhou, B., Jiang, Y., Sun, J.: Unified perceptual parsing for scene understanding. In: ECCV. pp. 418--434 (2018)

\bibitem{xie2021segformer}
Xie, E., Wang, W., Yu, Z., Anandkumar, A., Alvarez, J.M., Luo, P.: Segformer: Simple and efficient design for semantic segmentation with transformers. NIPS  \textbf{34},  12077--12090 (2021)

\bibitem{yan20222dpass}
Yan, X., Gao, J., Zheng, C., Zheng, C., Zhang, R., Cui, S., Li, Z.: 2dpass: 2d priors assisted semantic segmentation on lidar point clouds. In: ECCV. pp. 677--695. Springer (2022)

\bibitem{yang2022cross}
Yang, C., Zhou, H., An, Z., Jiang, X., Xu, Y., Zhang, Q.: Cross-image relational knowledge distillation for semantic segmentation. In: CVPR. pp. 12319--12328 (2022)

\bibitem{yu2018bisenet}
Yu, C., Wang, J., Peng, C., Gao, C., Yu, G., Sang, N.: Bisenet: Bilateral segmentation network for real-time semantic segmentation. In: ECCV. pp. 325--341 (2018)

\bibitem{yuan2020object}
Yuan, Y., Chen, X., Wang, J.: Object-contextual representations for semantic segmentation. In: ECCV. pp. 173--190. Springer (2020)

\bibitem{zhang2021video}
Zhang, J., Xu, X., Shen, F., Yao, Y., Shao, J., Zhu, X.: Video representation learning with graph contrastive augmentation. In: ACM MM. pp. 3043--3051 (2021)

\bibitem{zhang2019multi}
Zhang, L., Danelljan, M., Gonzalez-Garcia, A., Van De~Weijer, J., Shahbaz~Khan, F.: Multi-modal fusion for end-to-end rgb-t tracking. In: ICCVW. pp.~0--0 (2019)

\bibitem{zhang2022lgd}
Zhang, P., Kang, Z., Yang, T., Zhang, X., Zheng, N., Sun, J.: Lgd: label-guided self-distillation for object detection. In: AAAI. vol.~36, pp. 3309--3317 (2022)

\bibitem{zhao2017pyramid}
Zhao, H., Shi, J., Qi, X., Wang, X., Jia, J.: Pyramid scene parsing network. In: CVPR. pp. 2881--2890 (2017)

\bibitem{zheng2021rethinking}
Zheng, S., Lu, J., Zhao, H., Zhu, X., Luo, Z., Wang, Y., Fu, Y., Feng, J., Xiang, T., Torr, P.H., et~al.: Rethinking semantic segmentation from a sequence-to-sequence perspective with transformers. In: CVPR. pp. 6881--6890 (2021)

\bibitem{zhou2017scene}
Zhou, B., Zhao, H., Puig, X., Fidler, S., Barriuso, A., Torralba, A.: Scene parsing through ade20k dataset. In: CVPR. pp. 633--641 (2017)

\end{thebibliography}
\end{document}

% \title{Noised-labels Help Make a Strong Teacher: A Novel Knowledge Distillation \\ Approach for Semantic Segmentation} 

\title{Make a Strong Teacher with Label Assistance:\\A Novel Knowledge Distillation Approach \\ for Semantic Segmentation (Appendix)} 


\titlerunning{Label Assisted Distillation for Segmentation}

\author{Shoumeng Qiu\inst{1}\orcidlink{0000-0003-4475-2303} \and
Jie Chen\inst{1}\orcidlink{0000-0002-5625-5729} \and
Xinrun Li\inst{2}\orcidlink{0000-0002-2548-2187} \and
Ru Wan\inst{3}\orcidlink{0009-0008-8151-0059}\and
Xiangyang Xue\inst{1}\orcidlink{0000-0002-4897-9209}\and
Jian Pu\inst{1}*\orcidlink{0000-0002-0892-1213}}


\institute{Fudan University, Shanghai, China
\\
\email{smqiu21@m.fudan.edu.cn; \{chenji19,xyxue,jianpu\}@fudan.edu.cn}
\and Bosch Center for Artificial Intelligence, Germany \and
Mogo.ai Information and Technology Co., Ltd, Beijing, China\\
\email{wanru@zhidaoauto.com}
}

\maketitle

\section{Overview of the Supplementary}
In this supplementary file, we provide the following information, which we hope will shed deeper insight into our contributions:

\begin{itemize}

\item Training Details
%\item Generation of Noised-labels
\item Training Efficiency Analysis
\item Experiments with More Configurations
\item Experiments on Other Tasks
\item Experiments for Model Self-enhancement

\item Distillation with More Powerful Teacher
\item Difference between Output Distribution of Different Structure Models
\end{itemize}

For the reference in the manuscript in \emph{Simple to Complex Model Distillation}, please refer to section \ref{sec:mse}. For the reference in the manuscript in \emph{Comparison with Stronger Teacher}, please refer to section \ref{sec:ddm}. 
 
We first highlight our advantages: Different from previous teacher models in distillation, which are either complex or require additional modalities, and are difficult to obtain for high-performance student models, we propose a novel approach to simplify teachers, eliminate the need for extra modalities, and can be effectively applied to high-performance students. Comparisons are shown below.  

\begin{figure}[h]
\centering
\includegraphics[width=.85\linewidth]{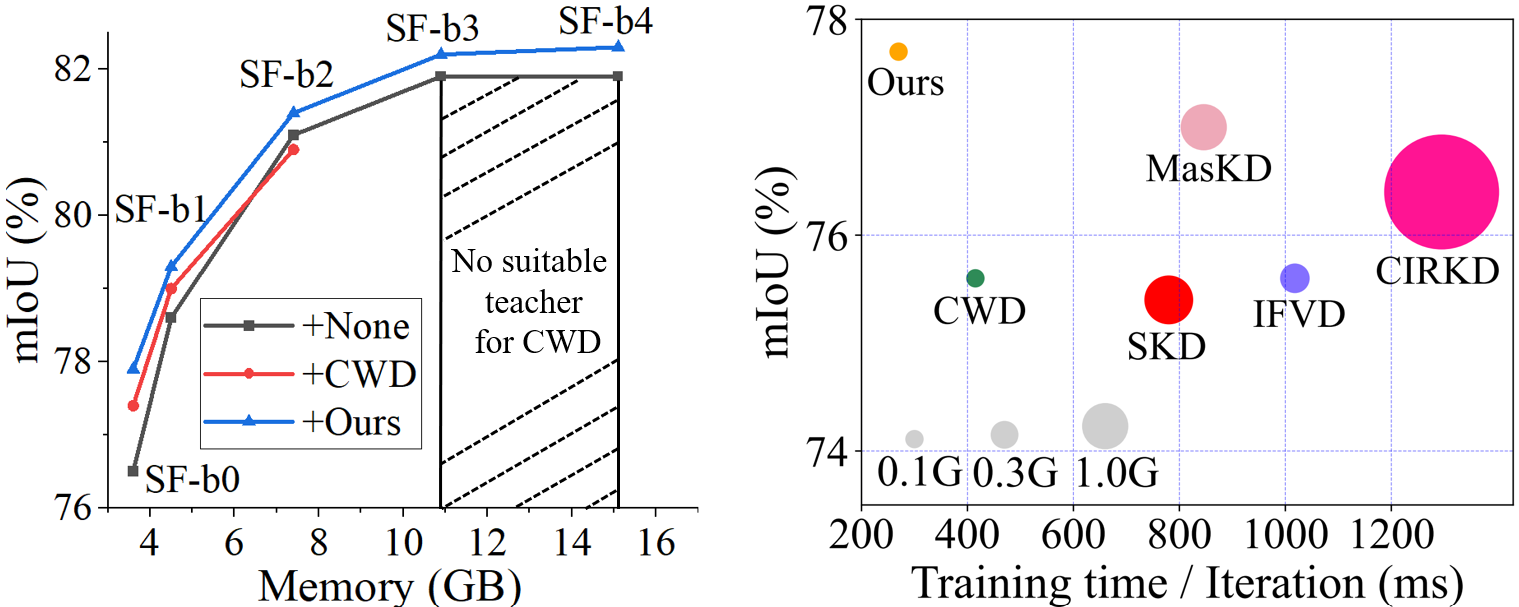}
\caption{Illustration of efficiency and effectiveness of our method. SF means SegFormer. Left: Teacher of CWD: SF-b4 (15.1G mem); our teacher (with noised label input): SF-b1(4.5G mem), memory usage is in inference. Right: point size represents FLOPs (G) of only distillation part. Model: DeepV3-R18, dataset: Cityscapes,  image size: 512$\times$1024.}
%\label{fig:tfd}
\end{figure}

\section{Training Details}
In our experiments, we utilize the codebase from \cite{mmseg2020} for most of the experiments in the manuscript, and we utilize the codebase from \cite{yang2022cross} for experiments in Table 3. This is because the codebase \cite{yang2022cross} provides the training code for the PSPNet-R18 and DeepLabV3-
Res18 model on the Pascal VOC dataset, which is not available in codebase \cite{mmseg2020}. For fair comparisons, we apply the same settings as the baseline methods (FCN, PSPNet, DeepLabV3, STDC, OCRNet). We adopt the official implementation code for the baseline distillation method CWD from \cite{shu2021channel}. For the sake of simplicity, we opt to solely employ knowledge distillation for the logits. During the teacher model training phase, the consistency loss coefficient is set to 1. For student model training, experimental settings from \cite{shu2021channel} was adopted, with the distillation temperature set to 4 and the distillation loss coefficient set to 3. For models with auxiliary heads,  the consistency loss and the distillation loss is only adopted on the primary task head. All experiments were conducted using 2 NVIDIA V100 GPUs. It should be noted that the official implementation of CWD has some differences in the experiment setting with ours (crop size). For a fair comparison, we also have compared our results with the CWD method with the same setting as ours in the codebase \cite{2021mmrazor}, which achieves 75.5\% for the PSPNet-R18 model and 76.6\% for the DeepLabV3-Res18 on the Cityscapes \cite{cordts2016cityscapes} val set in our implementation. Nevertheless, our approach still outperforms baselines. We summarize the training procedure of the teacher model in Alg. \ref{alg:search}.

\begin{algorithm}[t]  
  \caption{Teacher Model Training}  
  \label{alg:search}  
  \begin{algorithmic}[1] 
    \Require 
        DATASET: $D$; Initialized teacher model: $T$; Maximum epochs: MaxEpoch
    \Ensure  Trained model $T$. 
        \While{ epoch $<$ MaxEpoch } 
            \For{ image $X_i$, label $Y_i$ in $D$ }
                \State \textcolor{dark-gray}{//Process the label twice using the label noising module in Fig 3:}
                \State  NoisedLabel1 = LNM($Y_i$),  NoisedLabel2 = LNM($Y_i$);
                \State \textcolor{dark-gray}{//Concatenate the NoisedLabel on the RGB  channels of the image:}
                \State  input1 = cat($X_i$, NoisedLabel1), input2 = cat($X_i$, NoisedLabel2);
                \State \textcolor{darkgray}{//Duplicate the teacher model into two copies:}
                \State $T$1, $T$2 = Duplicate($T$);
                \State \textcolor{dark-gray}{//Forward propagate to obtain the prediction: $O_i^1$ and $O_i^2$}:
                \State $O_i^1$ = $T$1(input1), $O_i^2$ = $T$2(input2)
                \State \textcolor{dark-gray}{//Calculate the consistency loss between the two prediction}:
                \State $\mathcal{L}_{cosis} = Distance(O_i^1, O_i^2)$;
                \State \textcolor{dark-gray}{//Calculate the total losses}: 
                \State $\mathcal{L}_{total}$ = $\mathcal{L}_{seg}(O_i^1, Y_i)$ + $\mathcal{L}_{seg}(O_i^2, Y_i)$ + $\lambda$ $\mathcal{L}_{cosis}$;
                \State Backpropagate and update the model  parameters;
            \EndFor
            \State epoch = epoch + 1
        \EndWhile
   \State Return Trained model $T$.
  \end{algorithmic}  
\end{algorithm}

\section{Training Efficiency Analysis} To further demonstrate the training efficiency of our proposed approach, we compare the training time consumption between different distillation methods. The comparison results are presented in Table \ref{tab:timecomp}, including both the teacher training and distillation training times. It can be seen that there is a clear advantage of our method over other methods, which is primarily because our teacher model can be the same as the lightweight student model. 

\begin{table}[t]
    \centering
    \scriptsize
    \caption{Teacher training and distillation time comparison between different distillation methods. Experiments are conducted with NVIDIA V100 GPU on the Cityscapes val set, the image size is $512 \times 1024$. }  
    \setlength{\tabcolsep}{1.mm}{
    \begin{tabular}{c|c|c|c|c|c|c|c}
    \hline
       \multicolumn{2}{c|}{Methods (Deepv3-R101 $\rightarrow$ PSPNet-R18) } & SKD & IFVD & CWD & CIRKD & MasKD & Ours \\ \hline
    
         \multirow{4}{*}{\makecell[c]{Time/iteration \\ (ms)}} & Teacher training & \multicolumn{5}{c|}{868} & \textbf{213} \\ \cline {2 - 8} 
         & \makecell[c]{Distillation training \\ (Only teacher inference)} & \multicolumn{5}{c|}{257} & \textbf{39} \\ \cline {2 - 8} 
         & Distillation training (Total) & 758 &  944 & 402 & 1210 & 757 & \textbf{259} \\ \hline
    \end{tabular}
    }
    \label{tab:timecomp}
\end{table}

\section{Experiments with More Configurations}

We conducted experiments with SegFormer (SF) and Mask2Former (M2F) on Cityscapes, results are shown below. * denotes the model takes noised label as input, $-$ means we are unable to conduct experiments as it is difficult to find suitable high-performance teachers. We can see that our method shows consistent performance improvement.

\begin{table}[H]
    \centering
    \scriptsize
    \caption{Performance comparison on transformer-based methods.} 
    \setlength{\tabcolsep}{1.2mm}{
    \begin{tabular}{l|c|c|c|c|c|c|c}
    \hline
        Teacher (Base/Ours) & \multicolumn{5}{c|}{SF-b4/SF-b1* }  & \multicolumn{2}{c}{M2F-R101/M2F-R50*}  \\ \hline
        Students & SF-b0 & SF-b1 & SF-b2 & SF-b3 & SF-b4 &  M2F-R50 & M2F-SwinT \\ \hline
        +None  & 76.5 & 78.6 & 81.1 & 81.9 & 81.9 & 80.4 &  81.7  \\ \hline
        +CWD (Base)  & 77.4 & 79.0 & 80.9 & - & - &80.6  & - \\ \hline
        +Ours  & \textbf{77.9} & \textbf{79.3} & \textbf{81.4} & \textbf{82.2} & \textbf{82.3} & \textbf{80.9}  &  \textbf{82.7}  \\ \hline
    \end{tabular}
    }
    \label{tab:sotacomp}
\end{table}

We conducted experiments with PSPNet-R18 model on the ADE20K dataset. we followed CIRKD and MasKD for experiment comparison. Following your comments, we added experiments on ADE20K dataset in table below (Teacher of others: PSP-R101, our teacher: PSP-R18*, student: PSP-R18). Our method also demonstrates superiority.

\begin{table}[h]
    \centering
    \scriptsize
    \caption{Performance comparison on the ADE20K dataset.} 
    \setlength{\tabcolsep}{3.mm}{
    \begin{tabular}{l|c|c|c|c|c|c}
    \hline
        Metrics & None & SKD & IFVD & CWD  & CIRKD & Ours \\ \hline
        mIoU & 33.0 & 34.5 & 34.5 & 37.0 & 35.1 &  \textbf{37.3} \\ \hline
        mAcc & 42.6 & 44.3 & 44.3 & 46.3 & 45.4 & \textbf{47.6} \\ \hline
    \end{tabular}
    }
    \label{tab:sotacomp}
\end{table}

We conducted further experiments to see if the performance can be further improved with data augmentations as shown in \cite{pan2023semi}. Specifically, we adopt data augmentation for both teacher and student training (DeepV3-R18 on Cityscapes). We can see that with data augmentation, the performance can be further improved. Thanks for the excellent advice.

\begin{table}[H]
    \centering
    \scriptsize
    \caption{Performance comparison with data augmentation.} 
    \setlength{\tabcolsep}{3.5mm}{
    \begin{tabular}{l|c|c|c}
    \hline
         & Baseline & Teacher (ours) & Student \\ \hline
        None &  74.2 &  79.8 & 77.7  \\ \hline
        + data aug &  - &  \textbf{80.1} & \textbf{78.0} \\ \hline
    \end{tabular}
    }
    \label{tab:sotacomp}
\end{table}

We conducted experiments and made comparisons with more SOTA methods, including BPKD\cite{liu2024bpkd}. The experiments are conducted with PSPnet-R18 model on Cityscapes dataset. The results are shown in Table below \ref{tab:sotacompbpkd}. Our method also demonstrates superiority.

\begin{table}[H]
    \centering
    \scriptsize
    \caption{Performance comparison with more SOTA methods. To reduce the computational cost, the crop size is reduced to 512 × 512, and training schedule is 40k iterations.} 
    \setlength{\tabcolsep}{3.mm}{
    \begin{tabular}{l|c|c|c|c|c|c|c}
    \hline
        Metrics & None & SKD & IFVD & CWD  & CIRKD & BPKD & Ours \\ \hline
        mIoU & 69.0 & 69.3 & 71.1 & 74.3 & 72.2 & 75.9 &  \textbf{76.4} \\ \hline
        mAcc & 75.2 & 75.4 & 77.5 & 81.0 & 78.8 & 82.6 & \textbf{83.0} \\ \hline
    \end{tabular}
    }
    \label{tab:sotacompbpkd}
\end{table}

\section{Experiments on Other Tasks}

We conduct experiments on two different tasks: \textit{monocular depth estimation (DepthFormer \cite{li2023depthformer} on NYU-Depth-V2 dataset)} and \textit{learning with noise for segmentation (PSPNet-R18 on Cityscapes dataset)}, with the results presented in the Table \ref{tab:othertask}. For the monocular depth estimation task, we generate noise depth labels by randomly scaling the depth map and adding pixel-wise noise. For the learning with noise for segmentation task, we generate the noised labels by applying random dropblock \cite{ghiasi2018dropblock} to the ground truth labels. The results demonstrate that our method substantially improves performance on both tasks.

\begin{table}[h]
    \centering
    \scriptsize
    \caption{Experimental results on monocular depth estimation and learning with noise for segmentation tasks.}
    \setlength{\tabcolsep}{4.mm}{
    \begin{tabular}{c|c|c||c|c|c}
    \hline
        \multicolumn{3}{c||}{Monocular Depth Estimation (RMSE $\downarrow$)}  & \multicolumn{3}{c}{Learning with noise labels (mIoU $\uparrow$)}  \\ \hline
        Student & Teacher & Distillation & Student & Teacher & Distillation \\ \hline
        0.402 & 0.355 & 0.389 & 66.8 & 75.7 & 69.9  \\ \hline
    \end{tabular}
    \label{tab:othertask}}
\end{table}

\section{Experiments for Model Self-enhancement}
\label{sec:mse}
As the teacher can be the same as the student in our approach, so our approach can also be seen as a self-enhanced technique to further improve the performance of high-performance models. To verify the above hypothesis, we conduct experiments on methods UperNet \cite{xiao2018unified} and OCRNet \cite{yuan2020object}, and the results are shown in Table \ref{tab:semt}. The experimental results show that our method can be used to further improve the performance of some high-performance models.

\begin{table}[h]
    \centering
    \scriptsize
    \caption{Experiments for Model Self-enhancement on UperNet and OCRNet models.} 
    \setlength{\tabcolsep}{3.mm}{
    \begin{tabular}{c|c|c|c}
    \hline
        Model & None & \textbf{+} Privileged information & Distillation results \\ \hline
        UperNet \cite{xiao2018unified} & 78.7 & 81.5 & 79.6 (\textbf{+}0.9) \\ \hline
        OCRNet \cite{yuan2020object} & 80.6 & 85.2 & 81.3 (\textbf{+}0.7)\\ \hline
    \end{tabular}
    }
    \label{tab:semt}
\end{table}

\section{Distillation with More Powerful Teacher}
%(which is also consistent with the teacher model adopted by the baseline methods) 
To further explore the impact of the performance of the teacher model on the distillation performance of the student model, we also conduct experiments using the DeepLabV3-R101 with privileged information as the teacher model, and the results are shown in Table \ref{tab:maincomp}.

From Table \ref{tab:maincomp}, it can be observed is that although the performance of the DeepLabV3-R101 model is improved further after input with privileged information, achieving a performance of 83.2\% mIoU. However, the performance of the final student model did not see a corresponding improvement, and the `DeepLabV3-R101* -> PSPNet-R18' distillation performance even decreased. We believe this is because the performance of the teacher model is much higher and the gap between it and the student model is too large, which makes it more difficult to effectively transfer knowledge from teacher to student. which is also consistent with many previous studies \cite{cho2019efficacy,furlanello2018born,mirzadeh2020improved}.

\begin{table*}[h]
    %\footnotesize
    \scriptsize
    %\small
    \centering
    \caption{Further experiments with the more powerful teacher. The results are reported in terms of mIoU on Cityscapes val dataset. All models are pretrained on ImageNet. $*$ denotes the model takes the noised labels as privileged information.}
    \setlength{\tabcolsep}{1.mm}{
    \begin{tabular}{c|c|c|c}
    %\midrule
    \hline
        \multirow{2}{*}{Method} & \multirow{2}{*}{ \makecell[c]{Teacher (Params (M) \//  \\ FLOPs (G)  \// mIoU (\%))}} & \multirow{2}{*}{ \makecell[c]{Student  (Params (M) \// \\ FLOPs (G)  \// mIoU  (\%))}} & \multirow{2}{*}{ mIoU (\%)} \\ 
         & & &  \\

        %\midrule
        \hline
        \makecell[c]{SKD\cite{liu2020structured} (TPAMI 2020)} & \multirow{5}[0]{*}{\makecell[c]{ DeepLabV3-Res101 \\ (61.1M\//2371.7G\//78.07)} } & \multirow{5}[0]{*}{ \makecell[c]{PSPNet-R18 \\ (12.9M\//507.4G\//72.55)} }  & 73.3 \\ 
        \makecell[c]{IFVD\cite{wang2020intra}  (ECCV 2020)} &  &  & 73.7 \\ 
        \makecell[c]{CWD\cite{shu2021channel} (ICCV 2021)} &  &  & 74.4 \\ 
        \makecell[c]{CIRKD\cite{yang2022cross} (CVPR 2022)} &  &  & 74.7 \\ 
        \makecell[c]{MasKD\cite{huang2022masked} (ICLR 2023)} &  &  & 75.3 \\ 
        %\makecell[c]{DIST\cite{huang2022knowledge} (NIPS 2022)} &  &  & \textbf{76.3} \\ 
        %\midrule
        \hline
        \rowcolor{gray!15} Ours & \colorbox{gray!15}{\makecell[c]{DeepLabV3-Res101* \\ (61.1M\//2371.8G\//83.2)}} & \colorbox{gray!15}{\makecell[c]{PSPNet-R18 \\ (12.9M\//507.4G\//72.55)}} & 75.6 \\ 
        \hline
        \rowcolor{gray!15} Ours & \colorbox{gray!15}{\makecell[c]{PSPNet-R18*  \\ (12.9M\//507.5G\//79.7)}} & \colorbox{gray!15}{\makecell[c]{PSPNet-R18 \\ (12.9M\//507.4G\//72.55)}} & \textbf{76.2} \\ \midrule
        %\hline
        \hline
        \makecell[c]{SKD\cite{liu2020structured} (TPAMI 2020)} & \multirow{5}[0]{*}{\makecell[c]{ DeepLabV3-Res101 \\ (61.1M\//2371.7G\//78.07)} } & \multirow{5}[0]{*}{ \makecell[c]{DeepLabV3-Res18 \\ (13.6M\//572.0G\//74.21)} }  & 75.4 \\ 
        \makecell[c]{IFVD\cite{wang2020intra} (ECCV 2020)} &  &  & 75.6 \\ 
        \makecell[c]{CWD\cite{shu2021channel} (ICCV 2021)} &  &  & 75.6 \\ 
        \makecell[c]{CIRKD\cite{yang2022cross} (CVPR 2022)} &  &  & 76.4 \\ 
        \makecell[c]{MasKD\cite{huang2022masked} (ICLR 2023)} &  &  & 77.0 \\ 
        %\makecell[c]{DIST\cite{huang2022knowledge} (NIPS 2022)} &  &  & 77.1 \\ 
        %\midrule
        \hline
        \rowcolor{gray!15} Ours & \colorbox{gray!15}{\makecell[c]{DeepLabV3-Res101* \\ (61.1M\//2371.8G\//83.2)}} & \colorbox{gray!15}{\makecell[c]{DeepLabV3-Res18 \\ (13.6M\//572.0G\//74.21)}} & \textbf{77.7} \\ 
        \hline
        \rowcolor{gray!15} Ours & \colorbox{gray!15}{\makecell[c]{DeepLabV3-Res18* \\ (13.6M\//572.1G\//79.8)}} & \colorbox{gray!15}{\makecell[c]{DeepLabV3-Res18 \\ (13.6M\//572.0G\//74.21)}} & \textbf{77.7} \\ 
        %\midrule
        \hline
    \end{tabular}}
    \label{tab:maincomp}%
\end{table*}

\begin{table*}[h]
%\scriptsize
%\footnotesize
%\small
\tiny
  \centering
  \caption{The KL distance between the output of different models.}
  \setlength{\tabcolsep}{.4mm}{
    %\begin{tabular}{c|c|c|c|c|c|c|c|c|c|c|c|c|c|c|c|c|c|c|c|c}
    \begin{tabular}{c|ccccccccccccccccccc|c}
    \hline
    & \multicolumn{19}{c|}{KL Distance } & \\
    \hline
    Settings  & \rotatebox{90}{road} & \rotatebox{90}{sidewalk} & \rotatebox{90}{building}  &  \rotatebox{90}{wall} & \rotatebox{90}{fence}  & \rotatebox{90}{pole}  & \rotatebox{90}{traffic light} & \rotatebox{90}{traffic sign} & \rotatebox{90}{vegetation} &  \rotatebox{90}{terrain} & \rotatebox{90}{sky} & \rotatebox{90}{person}  & \rotatebox{90}{rider}  & \rotatebox{90}{car}   & \rotatebox{90}{truck}  & \rotatebox{90}{bus} & \rotatebox{90}{train}  & \rotatebox{90}{motorcycle} & \rotatebox{90}{bicycle} & \rotatebox{90}{mIoU (\%)} \\
    
    \hline
    % \makecell[c]{Deep101*->Deep18} & 0.02	& 0.12	& 0.05	& 0.53	& 0.40	& 0.22 & 0.19	& 0.19	& 0.05 & 0.31 & 0.02	& 0.11	& 0.21	& 0.03	& 0.30	& 0.14	& 1.28	& 0.31	& 0.15 & 77.7 \\
    % % \makecell[c]{Deep101->Deep18} & 0.01	& 0.07	& 0.03	& 0.33	& 0.25	& 0.11 & 0.12	& 0.14	& 0.02 & 0.16 & 0.01	& 0.08	& 0.21	& 0.02	& 0.21	& 0.13	& 0.27	& 0.29	& 0.11 & 75.6 \\
    % \makecell[c]{Deep101*->PSP18} & 0.02	& 0.18	& 0.06	& 0.50	& 0.51	& 0.24	& 0.19	& 0.18	& 0.05 & 0.35 & 0.03	& 0.11	& 0.30	& 0.04	& 0.53	& 0.25	& 1.26	& 0.43	& 0.18 & 75.6 \\
    % \hline
    % \makecell[c]{Deep101->Deep18} & 0.01	& 0.07	& 0.03	& 0.33	& 0.25	& 0.11 & 0.12	& 0.14	& 0.02 & 0.16 & 0.01	& 0.08	& 0.21	& 0.02	& 0.21	& 0.13	& 0.27	& 0.29	& 0.11 & 75.6 \\
    % \makecell[c]{Deep18*->Deep18} & 0.03	& 0.11	& 0.06	& 0.42	& 0.32	& 0.23	& 0.20	& 0.16	& 0.05 & 0.27 & 0.03	& 0.11	& 0.18	& 0.03	& 0.18	& 0.18	& 0.22	& 0.23	& 0.17 & 76.2 \\
    %\hline
    \makecell[c]{Deep101->stdc1} & 0.23  & 0.23  & 0.25  & 0.49  & 0.40   & 0.23  & 0.26  & 0.24  & 0.24 &  0.27 & 0.20   & 0.23  & 0.29  & 0.22  & 0.78  & 0.34  & 0.33  & 0.50   & 0.23 & 72.3 \\
    \makecell[c]{stdc1*->stdc1} & 0.07  & 0.13  & 0.11  & 0.53  & 0.48  & 0.19  & 0.21  & 0.17  & 0.09  & 0.33  & 0.04  & 0.15  & 0.23  & 0.09  & 0.50   & 0.22  & 0.31  & 0.30   & 0.18 & 73.1  \\
    \hline
    \end{tabular}}%
  \label{tab:cmvssm}%
\end{table*}%

\section{Difference between Output Distribution of Different Structure Models}
\label{sec:ddm}

Although lightweight models with privileged information can achieve comparable or better performance than complex models, and finally result in better performance of student models with distillation. A question arises: \emph{what are the differences between the outputs of power-pretrained teacher models and lightweight models trained with privileged information compared with the output of student model?} However, as the interpretability limitations of deep neural networks make theoretical analysis very challenging. Hence, we present experimental results and some discussions, hoping they will deepen our understanding of this problem.

Specifically, to understand the difference between teacher output and student output under different distillation settings, we calculate distribution distances for power-pretrained models to lightweight models with different structures and lightweight models with privileged information to lightweight models under the same structures. We adopt the Kullback-Leibler (KL) distance to measure the difference between the distribution of student and teacher models in each setting. The results are presented in Table \ref{tab:cmvssm}.

From Table \ref{tab:cmvssm}, we observe that the difference between the power-pretrained model and lightweight models with different structures is much higher than that between the lightweight model with privileged information (stdc*) and the lightweight model under the same structures. In addition, it is worth noting that a smaller distribution difference between the teacher and student models results in superior performance of the student model after knowledge distillation training. Specifically, for `stdc1*->stdc1', where the difference between the teacher and student models is minimal, the distillation result is 73.1\%. On the other hand, for the `Deep101->stdc1', the distillation results is noticeably lower (72.3\%) as the teacher-student model distribution difference is much larger compared to the distribution difference of `stdc1*->stdc1'.

%\clearpage

\bibliographystyle{splncs04}
\bibliography{main}